\documentclass[journal]{IEEEtran}

\usepackage{graphicx}
\usepackage{multirow}
\usepackage{array}
\usepackage{amsmath}
\usepackage{amssymb}
\usepackage{stmaryrd}
\usepackage{amsthm}
\usepackage{algorithm}
\usepackage{algpseudocode}
\usepackage{bm}
\usepackage{makecell}
\usepackage{subfig}
\usepackage{dblfloatfix}
\usepackage{color}
\usepackage{siunitx}

\newsavebox\CBox

\newtheorem{theorem}{Theorem}

\DeclareMathOperator*{\argmax}{arg\,max}

\algblockdefx[Foreach]{Foreach}{EndForeach}[1]{\textbf{foreach} #1 \textbf{do}}{\textbf{end foreach}}

\begin{document}
\title{Bayes Imbalance Impact Index: A Measure of Class Imbalanced Dataset for Classification Problem}

\author{Yang~Lu,~\IEEEmembership{Student Member,~IEEE,}
        Yiu-ming~Cheung,~\IEEEmembership{Fellow,~IEEE,}
        and~Yuan~Yan~Tang,~\IEEEmembership{Life Fellow,~IEEE}
\thanks{Yang Lu and Yiu-ming Cheung are with the Department of Computer Science, Hong Kong Baptist University, Hong Kong, China (email: \{yanglu, ymc\}@comp.hkbu.edu.hk). Yiu-ming Cheung is the corresponding author.}
\thanks{Yuan Yan Tang is with the Department of Computer and Information Science, Faculty of Science and Technology, University of Macau, Macau, China (email: yytang@umac.mo).} }


\maketitle


\begin{abstract}
Recent studies have shown that imbalance ratio is not the only cause of the performance loss of a classifier in imbalanced data classification. In fact, other data factors, such as small disjuncts, noises and overlapping, also play the roles in tandem with imbalance ratio, which makes the problem difficult. Thus far, the empirical studies have demonstrated the relationship between the imbalance ratio and other data factors only. To the best of our knowledge, there is no any measurement about the extent of influence of class imbalance on the classification performance of imbalanced data. Further, it is also unknown for a dataset which data factor is actually the main barrier for classification. In this paper, we focus on Bayes optimal classifier and study the influence of class imbalance from a theoretical perspective. Accordingly, we propose an instance measure called Individual Bayes Imbalance Impact Index ($IBI^3$) and a data measure called Bayes Imbalance Impact Index ($BI^3$). $IBI^3$ and $BI^3$ reflect the extent of influence purely by the factor of imbalance in terms of each minority class sample and the whole dataset, respectively. Therefore, $IBI^3$ can be used as an instance complexity measure of imbalance and $BI^3$ is a criterion to show the degree of how imbalance deteriorates the classification. As a result, we can therefore use $BI^3$ to judge whether it is worth using imbalance recovery methods like sampling or cost-sensitive methods to recover the performance loss of a classifier. The experiments show that $IBI^3$ is highly consistent with the increase of prediction score made by the imbalance recovery methods and $BI^3$ is highly consistent with the improvement of F1 score made by the imbalance recovery methods on both synthetic and real benchmark datasets.
\end{abstract}


\section{Introduction}

Classification of the binary imbalanced data is a challenging problem in the field of machine learning \cite{yang200610}. It refers to the problem that the classification accuracy is deteriorated when the number of samples in one class overwhelms another class. In this situation, even neglecting all the minority class samples can hardly effect the overall accuracy, because the minority class only takes a small percentage. This problem usually happens in detection tasks such as cancerous diagnosis \cite{rao2006data}, insider threat \cite{azaria2014behavioral} and software defect prediction \cite{wang2013using}, where the recognition target is the minority class that has relative small number of samples but draws more interests in the application domain. In the past decade, a number of imbalance recovery methods have been proposed. The objective of them is to improve the accuracy on the minority class without heavily sacrificing the accuracy on the majority class. A comprehensive review of the imbalance recovery methods can be found in \cite{he2009learning, branco2016survey}. These methods try to recover the performance loss caused by imbalance by virtue of preprocessing the training data or modifying the decision making procedure of an algorithm so that the minority class receives the same importance as the majority class during modeling and predicting.

\begin{figure*}[!tbp]
	\centering
	\subfloat[]{
	\includegraphics[width=.33\textwidth]{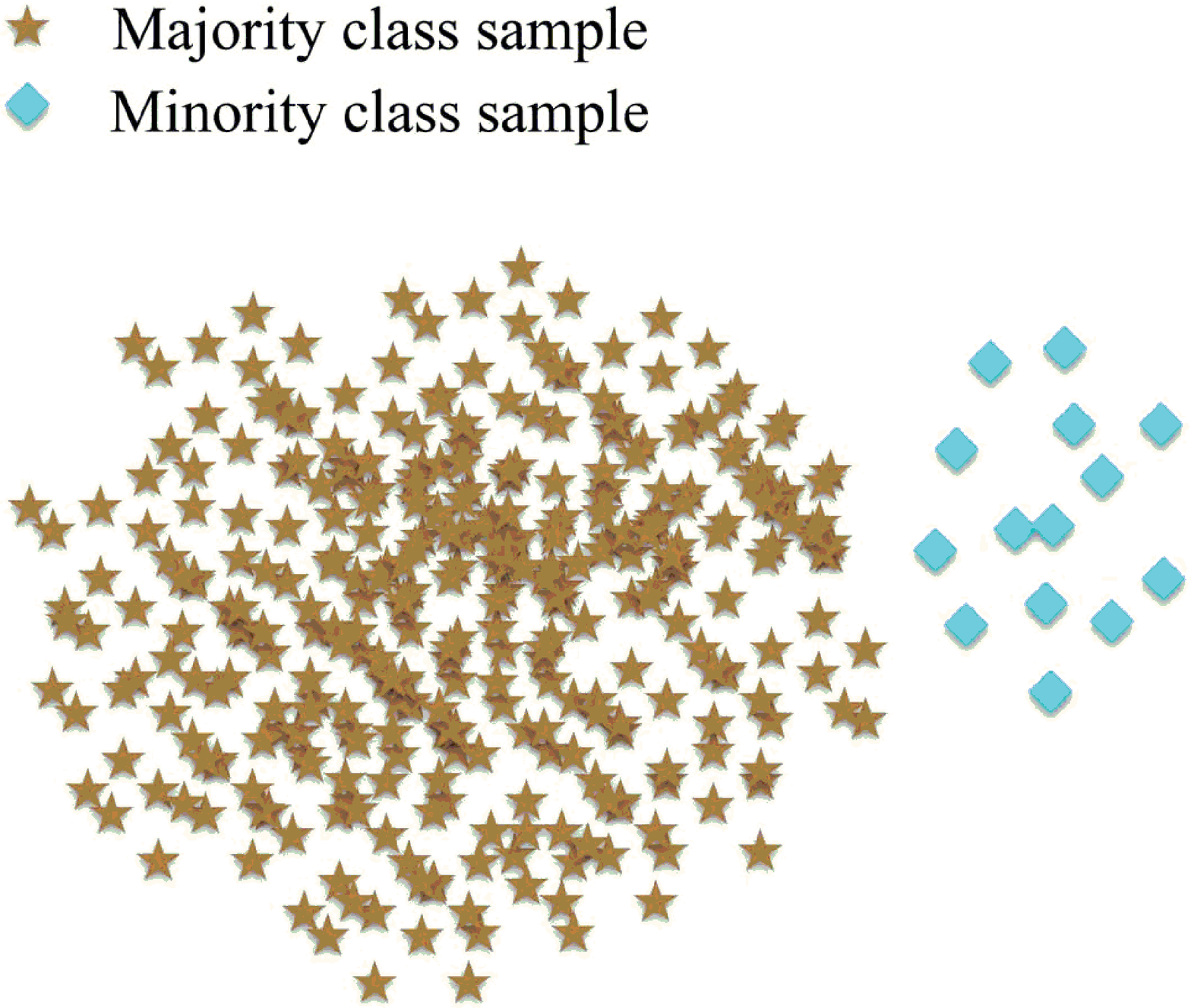}}
	\subfloat[]{
	\includegraphics[width=.33\textwidth]{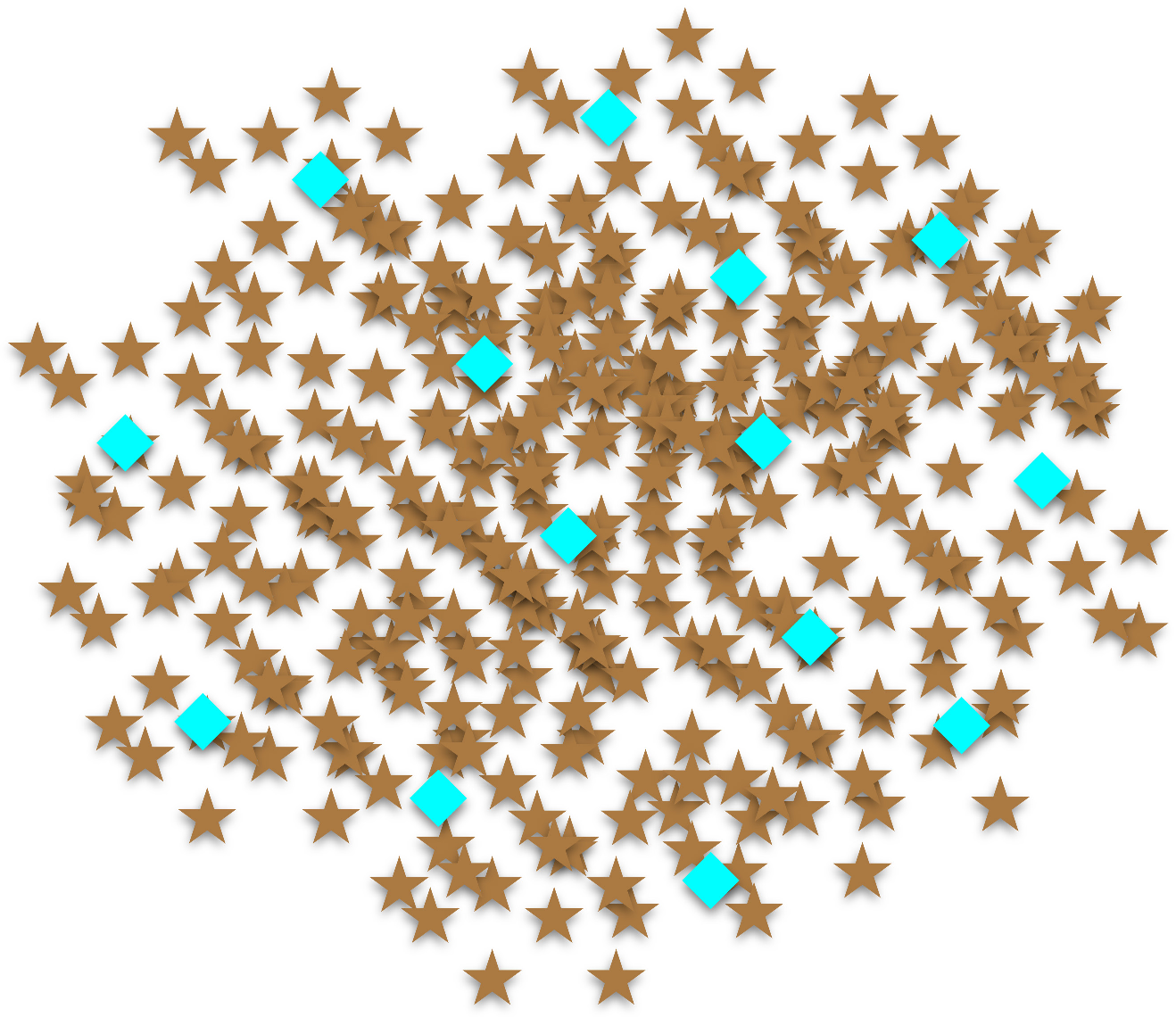}}
	\subfloat[]{
	\includegraphics[width=.33\textwidth]{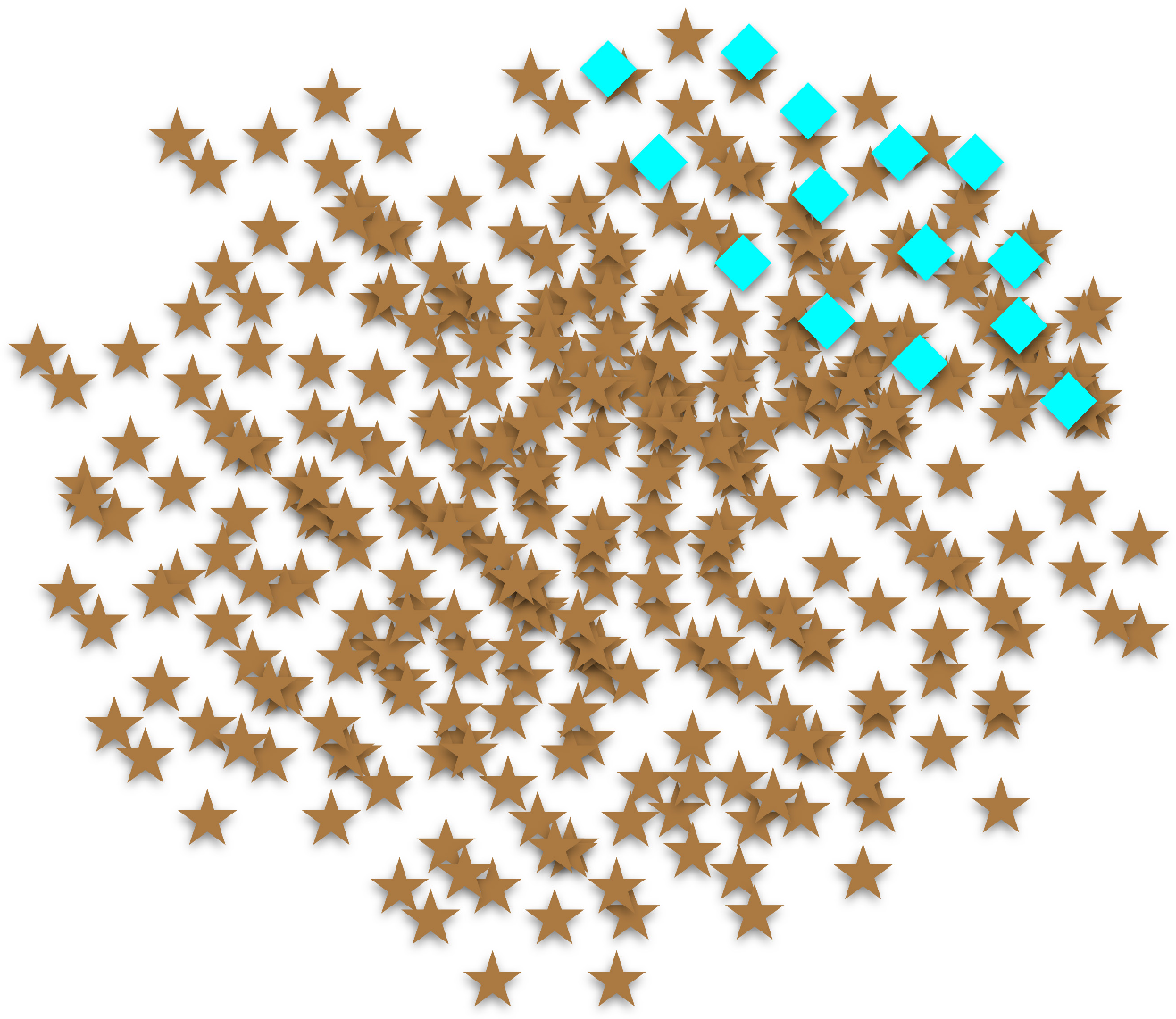}}
	\caption{Three imbalanced datasets with the same number of majority and the minority class samples. the minority class and the majority class are (a) separable, (b) totally overlapped, (c) partially overlapped.}
	\label{example1}
\end{figure*}

However, before adopting the imbalance recovery methods on an imbalanced dataset, one question should be raised first: Does one really have to take the so-called ``imbalanced'' issue into account using imbalanced recovery method, as given dataset that is more or less imbalanced? To answer this question, we should first define what kind of datasets are regarded as imbalanced, because the perfect balanced datasets are also very rare from the practical viewpoint. Usually, the researchers refer to the imbalance ratio (IR), which is the ratio between the number of the majority class samples and the minority class samples, to reflect the classification difficulty caused by class imbalance \cite{van2007experimental}. It has been commonly acknowledged that the higher IR, the more difficult to predict the minority class samples. However, recent studies have empirically shown that there is no obvious dependence between IR and the classification result \cite{lopez2013insight}. For example, Figure \ref{example1} shows three imbalanced datasets with the same IR. Actually, the accuracy improvement on the minority class from imbalance recovery methods on these three datasets are different. The two classes of the dataset shown in Figure \ref{example1}(a) are totally separated. In this case, no matter how severe the imbalance is, all samples will be correctly classified. On the contrary, the two classes of the dataset in Figure \ref{example1}(b) are totally and uniformly overlapped. Even imbalance recovery methods are applied, the best result is to recover at most half of the minority class samples in the cost of losing the accuracy of half of the majority class samples. For the case in Figure \ref{example1}(c), the minority class is partially overlapped with the majority class. If imbalance recovery methods are applied, most of the minority class samples can be correctly classified with the loss of a small amount of the majority class accuracy. In summary, if we only use IR to measure the difficulty of an imbalanced dataset, all three datasets in Figure \ref{example1} will be deemed to have the same difficulty for classification. Actually, the imbalance recovery methods cannot improve the classification of datasets in Figure \ref{example1}(a), and the extent of improvement is also different on datasets in Figure \ref{example1}(b) and (c). Therefore, if a dataset can hardly be improved by any imbalance recovery method, it is not necessary to consider the imbalance issue for this dataset. After all, sometimes the imbalance recovery methods may not only increase the computational burden, but also deteriorate the performance, if the cost of improving the minority class accuracy is to sacrifice more majority class accuracy.
It is also worth noting that IR is not the only factor that jeopardize the classification accuracy \cite{nathalie2003class,jo2004class}. Actually, the poor result can also be generated from both low IR and high IR. Therefore, other data factors should be considered as well when dealing with the imbalanced dataset. Basically, there are three data factors that are usually related to the class imbalance problem \cite{lopez2013insight}:

\begin{itemize}
  \item Small disjuncts: When the data in the same class is represented by different clusters, the underrepresented small cluster will further hamper the classification if imbalance exists in the dataset.
  \item Noise: The existence of noises in either the majority class or the minority class will bring extra difficulty, especially for the sampling-based imbalance recovery methods \cite{saez2015smote}.
  \item Overlapping: The degree of overlapping highly effects the minority class accuracy because sacrificing the minority class samples in the overlapping region usually get higher overall accuracy in return.
\end{itemize}

Currently, most of the existing work empirically analyzes the relationship between the three data factors and imbalance by experiments. To the best of our knowledge, no theoretical analysis on such relationship has been conducted thus far. Instead, the only conclusion is that, under the same degree of other data factors such as overlapping, small disjunct and noise, higher IR may further deteriorate the performance \cite{nathalie2003class,jo2004class}. However, the data factors are different for different datasets. Purely using IR to represent the difficulty of the imbalanced dataset is insufficient and inaccurate. In other words, given an imbalanced dataset with low performance, one has no idea whether this performance loss is due to the imbalance or other factors. To obtain the degree of imbalance impact by isolating other data factors and fill the gap of the research problem, this paper therefore proposes two new measures called Individual Bayes Imbalance Impact Index ($IBI^3$) and Bayes Imbalance Impact Index ($BI^3$) to estimate the degree of deterioration caused purely by imbalance on instance level and data level, respectively. $IBI^3$ is calculated by quantizing the difference of prediction score of a given minority class sample between the imbalanced and balanced situation. $BI^3$ is the averaged $IBI^3$ over all minority class samples and can therefore be used to describe the imbalance impact to the dataset. Back to the previous example, the dataset in Figure \ref{example1}(a) will have very small $BI^3$ and the one in Figure \ref{example1}(c) will have larger $BI^3$ than the one in Figure \ref{example1}(b). Therefore, $BI^3$ can be used as a judgement index, instead of purely referring to IR, to determine whether we should consider the imbalance issue and whether imbalance recovery methods should be applied before training the dataset. That is, $BI^3$ has positive correlation with the benefit of applying imbalance recovery methods. The higher $BI^3$ is, the more performance improvement can be made by imbalance recovery methods. We conduct the experiments to verify the effectiveness of $IBI^3$ and $BI^3$ by correlation analysis with the different standard classifiers and different imbalance recovery methods. Experimental results show that $IBI^3$ has high correlation with the increase of prediction score on minority class samples, and $BI^3$ has high correlation with the improvement of F1 score on the whole data on both synthetic and real benchmark datasets. Therefore, $BI^3$ is a suitable measure to describe how the data is influenced by imbalance. The contribution of this paper is summarized as follows:
\begin{itemize}
	\item This paper is the first attempt to study the data factors of imbalanced dataset from a theoretic perspective.
	\item The proposed $IBI^3$ is the first instance complexity to show how a minority class sample is influenced by imbalance.
	\item The proposed $BI^3$ can be used as a data complexity measure to describe the imbalance degree, instead of only referring to IR.
	\item The influence of the imbalance can be estimated without training and testing, so that one can determine whether to apply a specific imbalance recovery method.
\end{itemize}

The rest of the paper is organized as follows. Section II lists the related work on class imbalance problem, and discusses the data factors related to imbalance problem. Section III describes the proposed method. Section IV presents the experiments and discussions. Finally, concluding remarks are given in Section V.

\section{Related Work}

Most of the existing work on class imbalance learning is to propose imbalance recovery methods. They can be basically categorized into three groups \cite{galar2012review}. The first group is on data level. The methods in this group aim to manipulate the data to be balanced before training. The most well-known method in this group is Synthetic Minority Over-sampling TEchnique (SMOTE) \cite{chawla2002smote}. It synthesizes new samples to the minority class by interpolating the minority class samples with their neighbors. In addition to data synthesis, data cleaning techniques have also been used in data preprocessing. For example, Batista \textit{et al.} \cite{batista2004study} adopted Tomek links to clean the overlapping area between classes so that the classification boundary becomes clear after introducing synthetic  samples. The second group is on algorithm level. They modify the existing learning methods by adapting them to the imbalanced data. The modified algorithm usually shift the decision bound to enhance the existence of the minority class samples. For example, Hong \textit{et al.} \cite{hong2007kernel} modified the kernel classifiers by orthogonal forward selection to optimize the model generalization for imbalanced datasets.  The last group is related to the framework of cost-sensitive learning \cite{elkan2001foundations}. They assign different costs to the samples in difference classes. Usually the minority class samples are assigned with a large cost so that they will not be easily misclassified. The idea of cost-sensitive can also be applied to many existing algorithms to turn them into imbalance recovery methods, such as decision tree \cite{ling2006test} and SVM \cite{davenport2010tuning}.

The imbalance recovery methods mentioned above assume the deteriorated performance is caused by the existence of class imbalance. Recent studies have shown that the imbalance is not the only cause for the performance deterioration \cite{lopez2013insight,saez2015smote,napierala2016types}. Actually, there are at least three other factors to make the prediction inaccurate on imbalanced datasets. The first factor is the sparsity of the minority class, where the minority class samples are separated into small clusters. This problem is called small disjuncts or within-class imbalance \cite{he2009learning}, which is often studied in tandem with the imbalance. Therefore, Japkowicz et al. \cite{japkowicz2003class} generated synthetic data to study the relationship among the class disjuncts, the size of the training data, and the imbalance ratio. The results show that the small disjuncts take more responsibility for the decrease in accuracy than the imbalance ratio by changing the degrees of these data factors. Accordingly, a solution dealing with small disjuncts called CBO has been proposed in \cite{jo2004class}. It conducts clustering on each class first so that the oversampling is conducted on each disjunct instead of each class. Besides, Prati et al. \cite{prati2004learning} studied the performance of unpruned trees by considering the relation between class imbalance and small disjuncts and proposed to use SMOTE with data cleaning methods to alleviate the performance loss from small disjuncts.

The second data factor is noise. Noisy samples are usually defined as the ones from one class located deep into the other class \cite{kubat1997addressing}. The existence of noise samples in the minority class will make blind oversampling methods like SMOTE generate more noises, so that applying oversampling on the noisy the minority class may even degrade the performance \cite{saez2015smote}. Therefore, data cleaning methods are usually adopted to tackle the noises such as Tomek link \cite{batista2004study} and ENN \cite{laurikkala2001improving} . Another straightforward method to find noise is to collect the samples which are wrongly classified by $k$NN classifier \cite{napierala2010learning}. Van Hulse and Khoshgoftaar experimented on data with artificial noises \cite{van2007experimental}, where the class noise is injected to real datasets by randomly relabelling the samples before training. The results show that the minority class is severely effected by noises with all compared classifiers.

The last factor is the overlapping between classes which effects classification, especially when the data is imbalanced. Napierala and Stefanowski \cite{napierala2016types} proposed a $k$NN-based method to category the minority class examples into 4 groups: safe, border, rare and outlier. The categories of 4 groups depend on the ratio of the majority class samples in the $k$ nearest neighbors of each minority class sample. For each dataset, the overlapping degree of the minority class can be obtained by investigating the portions of the 4 groups. However, the analysis only shows the difficulty of classifying the minority class samples. The degree of imbalance is not considered. Garc{\'\i}a et al. \cite{garcia2008k} evaluated $k$NN in the situation that the local imbalance ratio is inverse to the global imbalance ratio and concluded that $k$NN is more dependent on the local imbalance. Recently, Anwar et al. \cite{anwar2014measurement} have also proposed to use $k$NN to measure the data complexity for imbalanced data with adaptively selected $k$. Prati et al. \cite{prati2004class} observed that the performance loss is not only related to class imbalance, but also the overlapping degree. To sum up, the existing work mentioned above all empirically justify their conjecture without a theoretical framework. In fact, they have yet to give a measure to assess how the dataset is influenced by class imbalance independent of other data factors.

Before we close this section, we would like to point out that another somewhat related area is data complexity. A list of complexity measures are proposed in \cite{ho2002complexity} with different featured groups. The measures are used to study the essential structure of data and guide classifier selection for specific problems. Recently, Smith et. al \cite{smith2014instance} have extented the data complexity from data level to instance level. They proposed a group of complexity measures that can be calculated for each instance. The correlation among those measures are then analyzed. The instance level complexity measures can be used for data cleaning that filters the most difficult samples in the data. However, there is no specific research on the data complexity for imbalanced data, and the existing complexity measures are not suitable to describe in what extent that the data is influenced by imbalance.

\section{Proposed Method}

In order to get the influence of imbalance on a dataset, a straightforward way is  to compare the model learned from the imbalanced data with the model learned from its balanced case, where the minority class samples with equal number of the majority class are drawn from the underlying distribution. If the distribution is known, it can be clearly figured out that how different are the models built on the imbalanced and balanced data, because other data factors fixed. However, the distribution is usually unknown from practical viewpoint. We can only estimate the distribution by the existing minority class samples in the dataset. Therefore, we propose to estimate the difference in light of Bayes optimal classifier, because it has the theoretical minimum classification error and the class prior is taken into account. Based on the Bayes decision theory, one can estimate the difference of the theoretical classification error between the classifiers trained on the imbalanced and balanced dataset. Thus, the impact of imbalance can be estimated while isolating other data factors which may influence the classification. First we decompose the problem into the instance level and propose Individual Bayes Imbalance Impact Index ($IBI^3$). It measures how each minority class sample is influenced during classification by class imbalance. Then, we define the data level measure as Bayes Imbalance Impact Index ($BI^3$), by averaging $IBI^3$ over all minority class samples. $BI^3$ thus represents the impact brought by imbalance on the whole data.

The details of the proposed measures are described as follows. By Bayes rule, the posterior probability of a given sample $\textbf{x}$ in class $c$ is
\begin{align}\nonumber
p(y=c|\textbf{x})=\frac{p(\textbf{x}|y=c)p(y=c)}{p(\textbf{x})}.
\end{align}
The decision of the optimal Bayes classifier for binary classification problem follows:
\begin{align}\nonumber
f(\textbf{x})=\argmax_{c=\{+1,-1\}}p(y=c|\textbf{x}).
\end{align}
Because $p(\textbf{x})$ is same for both classes and in practice the prior probability is usually estimated by the frequency of each class. The decision can then be formulated as:
\begin{align}\nonumber
f(\textbf{x})=\left\{\begin{array}{cl}+1,& f_p(\textbf{x})>f_n(\textbf{x}),\\-1,& \mathrm{otherwise},\end{array}\right.
\end{align}
where
\begin{align}\nonumber
f_p(\textbf{x})&=N_pp(\textbf{x}|+),\\\nonumber
f_n(\textbf{x})&=N_np(\textbf{x}|-),
\end{align}
and $N_p$ and $N_n$ are the number of samples in the positive class and negative class respectively and $f_p(\textbf{x})$ and $f_n(\textbf{x})$ are the posterior scores which are proportional to the posterior probabilities. $y=+1$ and $y=-1$ are simplified as $+$ and $-$ in the conditional probability. Usually, we denote the majority class as negative and the minority class as positive. When the class is imbalanced, namely $N_p<N_n$, the Bayes optimal decision may be dominated by the frequency such that some or even all minority class samples may be misclassified. Because the optimal Bayes error is the sum of all misclassified samples regardless of the class, under the imbalance circumstance, sacrificing the accuracy of the minority class samples helps minimize the total error. However, in most of the imbalanced data applications, low error rate does not represent good performance. The minority class is usually more important and F1, G-mean and AUC are the common used measurements instead of error rate \cite{he2009learning}. Thus, the alternative decision function that is not influenced by the prior probability can be written as:
\begin{align}\nonumber
f'(\textbf{x})=\left\{\begin{array}{cl}+1,& f'_p(\textbf{x})>f_n(\textbf{x}),\\-1,& f'_p(\textbf{x})<f_n(\textbf{x}),\end{array}\right.
\end{align}
where
\begin{align}\nonumber
f'_p(\textbf{x})&=N_nf(\textbf{x}|+).\\\nonumber
\end{align}
The decision function $f'(\textbf{x})$ directly compares the value between $p(\textbf{x}|+)$ and $p(\textbf{x}|-)$. It is actually the decision function with minimal Bayes error when the classes are balanced. The influence of imbalance on the dataset can be reflected by the difference between $f_p'$ and $f_p$, where $f_p$ is proportional to the minority class posterior probability under the real imbalanced case and $f_p'$ is under the estimated balanced case. However, directly comparing $f_p$ and $f_p'$ is meaningless because the decision hyperplane is also determined by $f_n$. Therefore, we define $IBI^3$ as the difference between normalized posterior probabilities between the imbalanced case and the estimated balanced case:

\begin{figure}[!t]
	\centering
	\subfloat[]{
	\includegraphics[width=.45\textwidth]{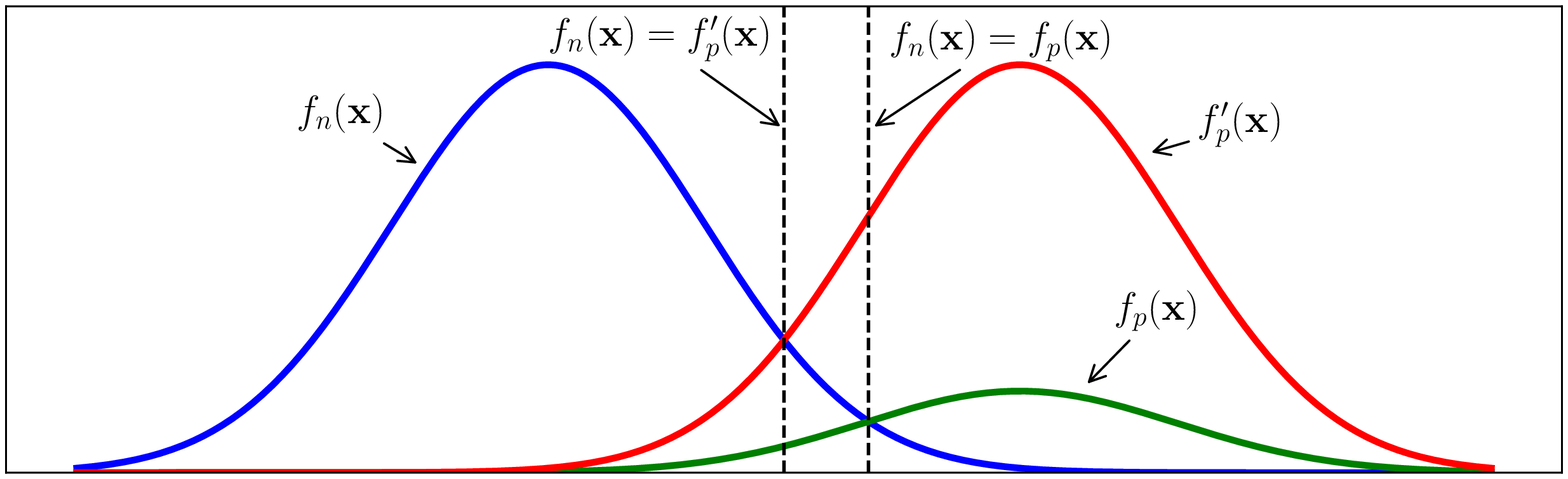}}\\
	\subfloat[]{
	\includegraphics[width=.45\textwidth]{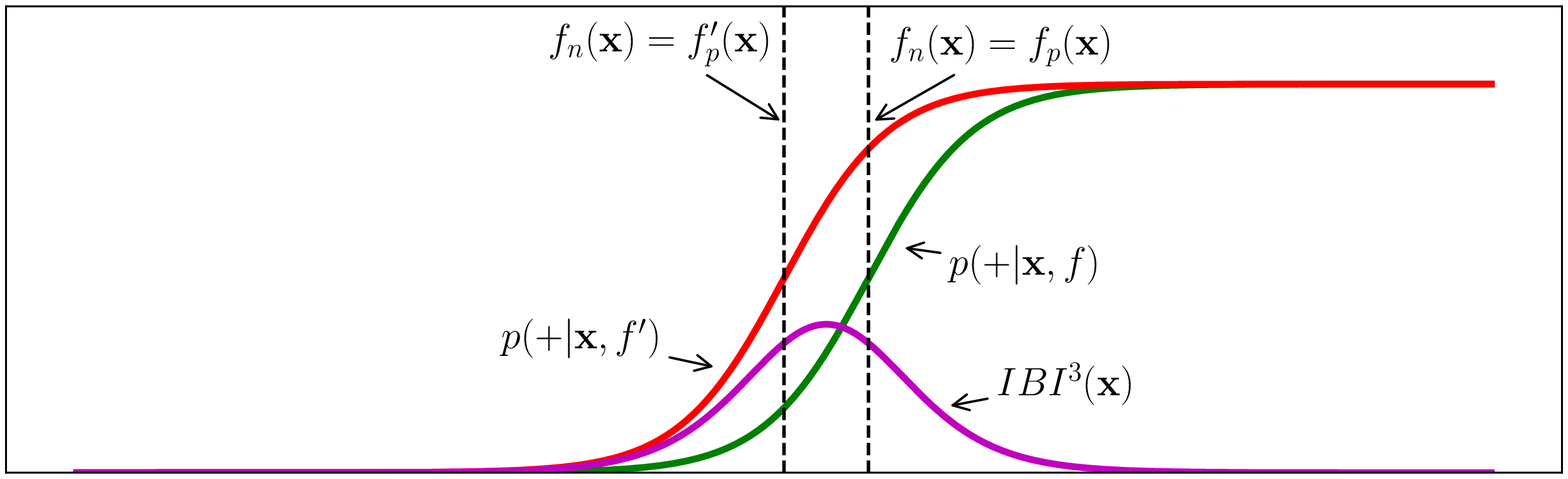}}
	\caption{An example to show the distribution of $IBI^3$ on two classes with normal distributions. (a) The posterior scores. (b) Normalized posterior probabilities and $IBI^3$. The optimal Bayes decision hyperplanes $f'(\textbf{x})$ and $f(\textbf{x})$ are shown by dotted lines.}
	\label{normal}
\end{figure}

\begin{align}\label{recovery_probability}
IBI^3(\textbf{x})&=p(+|\textbf{x},f')-p(+|\textbf{x},f)\\\label{recovery_probability2}
&=\frac{f'_p(\textbf{x})}{f_n(\textbf{x})+f'_p(\textbf{x})}-\frac{f_p(\textbf{x})}{f_n(\textbf{x})+f_p(\textbf{x})}.
\end{align}
Figure \ref{normal}(a) shows an example of the distribution of $f_n(\textbf{x})$, $f_p(\textbf{x})$ and $f_p'(\textbf{x})$ on an one dimensional normally distributed binary class data with $IR=5$. Figure \ref{normal}(b) shows the normalized posterior probabilities and $IBI^3$. It can be observed that the peak of $IBI^3$ locates in the region between two decision hyperplanes $f(\textbf{x})$ and $f'(\textbf{x})$, which means that the most difference part between the imbalanced and balanced case is in the region between two hyperplanes. The minority class samples in this region is misclassified under the imbalanced case but correctly classified under the balanced case, which can be regarded as the impact to the minority class sample solely from the imbalance. If $IBI^3$ is low, the minority class sample $\textbf{x}$ is either a noise sample, which is deeply located in the region of the majority class that makes both $p(+|\textbf{x},f')$ and $p(+|\textbf{x},f)$ close to 0, or a safe sample which is deeply located in the region of the minority class that makes both $(p(+|\textbf{x},f')$ and $p(+|\textbf{x},f))$ close to 1. In both cases, $IBI^3$ is small and such $\textbf{x}$ is hardly influenced by the imbalance.

\begin{algorithm}[!t]
\caption{Bayes Imbalance Impact Index}
\label{alg1}
\begin{algorithmic}[1]
	\Require Dataset $\mathcal{D}=\{\mathbf{x}_i\in\mathcal{X},y_i\in\mathcal{Y}\}$, the number of positive samples $N_p$, the number of negative samples $N_n$, the number of nearest neighbors $k_0$.
	\State $r=N_n/N_p$;
	\State Construct the set of all the minority class samples $\mathcal{D}^+=\{\textbf{x}^+_i\}$;
	\For{$i\gets 1$ to $N_p$}
		\State Calculate the number of the majority class neighbors:
		\Statex ~~~~$M=|\{(\textbf{x}',y'):\textbf{x}'\in kNN(\textbf{x}^+_i), y'=-1\}|$
		\If{$M=0$}
			\State $M\gets$ the number of the majority class samples
			\Statex ~~~~~~~~~~~~~~~~between $\textbf{x}^+_i$ and the nearest the minority class
			\Statex ~~~~~~~~~~~~~~~~neighbor of $\textbf{x}^+_i$;
			\State $k=M+1$;
		\Else
			\State $k=k_0$;
		\EndIf
		\State $f_n\gets M/k$;
		\State $f_p\gets (k-M)/k$;
		\State $f'_p\gets r(k-M)/k$;
		\State Calculate $IBI^3(\textbf{x}^+_i)$ by (\ref{recovery_probability2});
	\EndFor
	\State Calculate $BI^3$ by (\ref{BI3});
	\Ensure The indices $IBI^3$ and $BI^3$.
	\end{algorithmic}
\end{algorithm}

$IBI^3$ is calculated for each minority class sample and the averaged $IBI^3$ over all the minority class can be used to describe the imbalance impact of the dataset. $BI^3$ for the whole dataset $\mathcal{D}$ is calculated by averaging over all $IBI^3$ on the minority class:
\begin{align}\label{BI3}
BI^3(\mathcal{D}) = \frac{1}{N_p}\sum_{\substack{(\textbf{x}_i,y_i)\in\mathcal{D},\\y_i=+1}}IBI^3(\textbf{x}_i).
\end{align}
If the two classes are normal distributed, the likelihood functions $p(\textbf{x}|+)$ and $p(\textbf{x}|-)$ can be calculated by estimating the mean and variance. However, the assumption usually fails in real benchmark datasets. Because not only the distribution is not normal, but also there are small disjuncts and noises among the classes. Suppose the normality with estimated mean and variance may not be accurate enough to calculate $IBI^3$ and $BI^3$. Cover and Hart \cite{cover1967nearest} have shown the relation between the error bounds of nearest neighbor classifier and Bayes classifier by the following theorem.
\begin{theorem}[Cover and Hart, 1967]
For sufficiently large training set size $N$, the inequality of the error rate of nearest neighbor classifier $R_{NN}$ and Bayes classifier $R_{Bayes}$ holds:
\begin{align}\nonumber
R_{Bayes}\le R_{NN}\le2R_{Bayes}(1-R_{Bayes}).
\end{align}
\end{theorem}
\noindent It has been shown that the upper bound of the error rate of nearest neighbor classifier is double of the error rate of Bayes classifier and the result is independent of the selection of $k$ for nearest neighbor. Therefore, $k$ nearest neighbors ($kNN$) is a good substitute to estimate the likelihood without normality assumption. The algorithm is shown in Algorithm \ref{alg1}. For each minority class sample $\textbf{x}$, we find its $k$ nearest neighbors $kNN(\textbf{x})$ and count the number of the majority class neighbors $M$. Thus, $f_n$ is set at $M/k$, which is the local probability that $\textbf{x}$ is classified as negative, and $f_p$ is set at $(k-M)/k$. We assume that in the unknown balanced situation, there will be $r=N_n/N_p$ times more the minority class samples surrounded by $\textbf{x}$. Therefore, $f'_p$ is set at $r(k-M)/k$. To prevent the case that all of the $k$ neighbors of $\textbf{x}$ are the majority class samples, which makes both $f_p$ and $f'_p$ equal to zero, we adopt a flexible $k$ that is set at the minimal number to make $\textbf{x}$ has at least one the minority class neighbor. It is shown in Line 5-10 in Algorithm \ref{alg1}.

\begin{figure}[!t]
	
	\subfloat{\hspace{.5cm}
	\includegraphics[width=.2\textwidth]{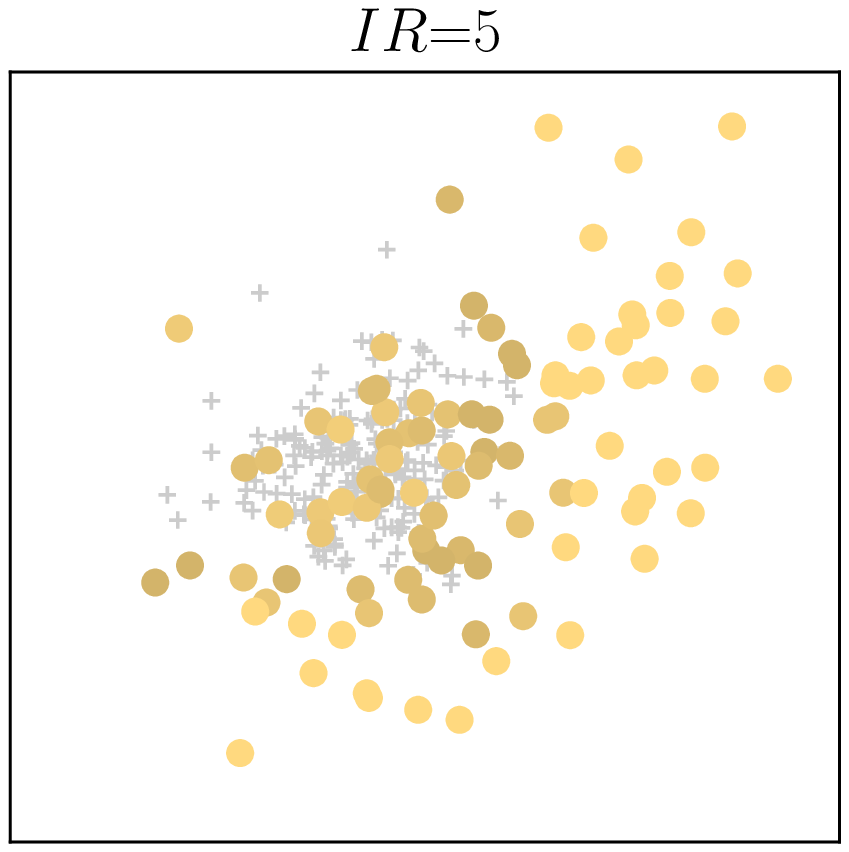}}
	\hspace{-.5cm}
	\subfloat{
	\includegraphics[width=.2\textwidth]{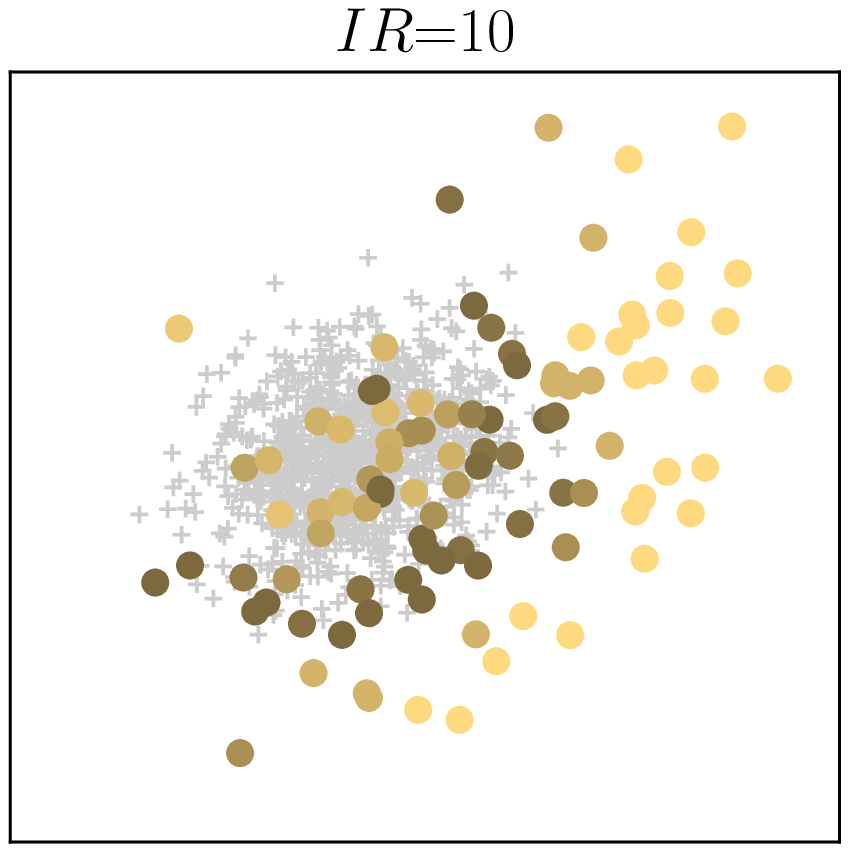}}\\
	\subfloat{\hspace{.5cm}
	\includegraphics[width=.2\textwidth]{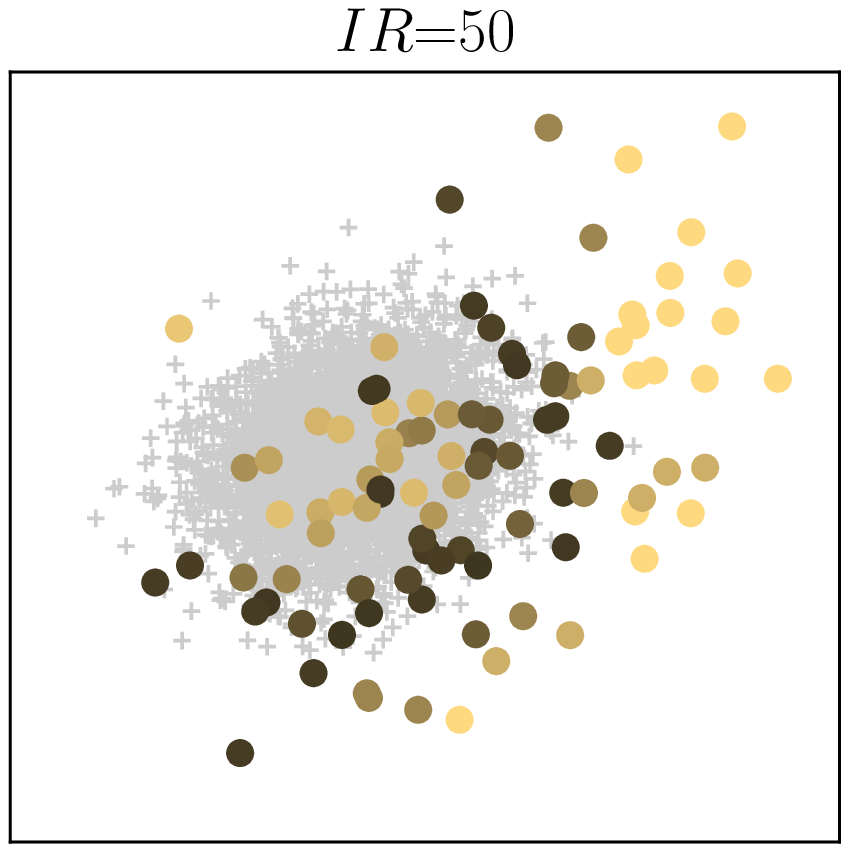}}
	\hspace{-.6cm}
	\subfloat{
	\includegraphics[width=.25\textwidth]{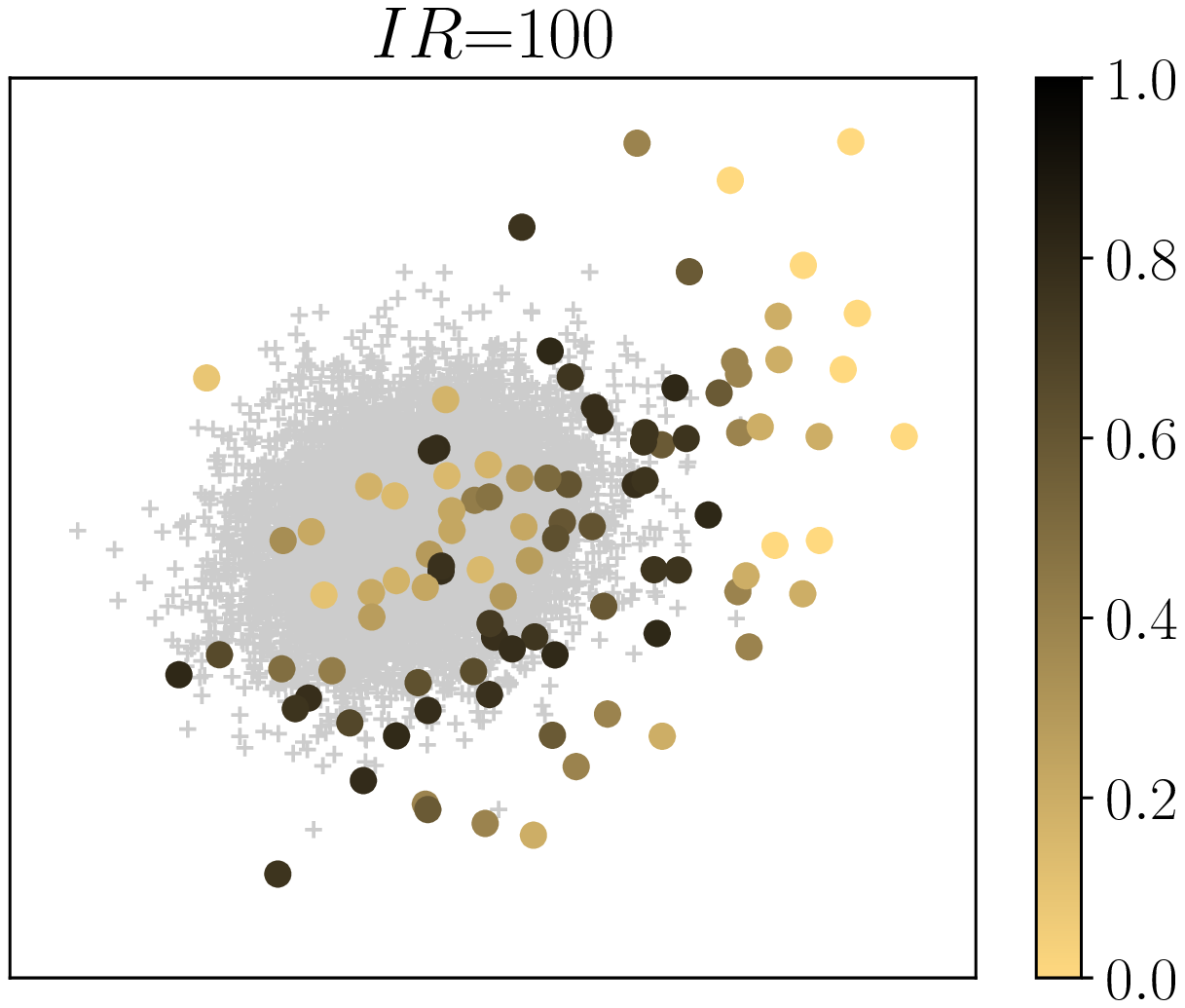}}
	\caption{Values of $IBI^3$ with local probability on a binary class synthetic dataset drawn from normal distribution with different imbalance ratios. The grey plus is the majority class and the colored circle is the minority class.}
	\label{colorbar_knn}
\end{figure}

An example with four binary class synthetic datasets drawn from normal distribution with different imbalance ratios is shown in Figure \ref{colorbar_knn}. The value of $IBI^3$ with $k_0=5$ can be visually compared with different locations of the minority class samples and with different $IR$. It can be observed that in Figure \ref{colorbar_knn}, the minority class samples with high values of $IBI^3$ mainly locate in the boundary between two classes. This is consistent with the example shown in Figure \ref{normal}. The minority class samples that are in the deep region of the majority class receives low $IBI^3$, because they are regarded as noises that will still be misclassified even if the two classes are balanced. Thus, the classification result of them is hardly related to the imbalance. In addition, the minority class samples that are far away from the majority class also receive low $IBI^3$, because they will be correctly classifier no matter the classes are imbalanced or not. From Figure \ref{colorbar_knn}(a) to (d), it can be observed that the value of $IBI^3$ of the minority class samples on the boundary between two classes increases as $IR$ increases. That means the influence of those the minority class samples are related to $IR$. The higher the value of $IBI^3$ of a minority class sample is, the more seriously that the sample is influenced by imbalance and the higher probability that the sample can be correctly classified in the balanced situation. The values of $BI^3$ of this four datasets are 0.0674, 0.2482, 0.3829 and 0.4588, respectively. The values of $BI^3$ increases as $IR$ increases and it can be used to reflect the extent that imbalance influences the data.

\section{Experiments}

In the experiments, the accuracy of the proposed measure $BI^3$ is evaluated by correlation analysis. We adopt Spearman's rank correlation coefficient \cite{kendall1990rank}, which is a nonparametric measure of rank correlation between two variables. It assesses the degree of describing the relationship between two variables by using a monotonic function. The correlation ranges from -1 to 1, where 1 or -1 indicates a perfect monotonously increasing or decreasing relationship and 0 indicates no correlation between two variables.

We adopt five well-known standard classifiers: RBF kernel Support Vector Machine (SVM) \cite{vapnik1998statistical}, Decision Tree implemented by CART \cite{breiman1984classification}, $k$ Nearest Neighbors with $k=5$ (5NN) \cite{duda1973pattern}, Random Forest (RF) \cite{breiman2001random} and AdaBoost \cite{freund1997decision}. We use the default parameter provided by $scikit$-$learn$ learning library in Python \cite{scikit-learn}. The minimal number of nodes in each leaf of CART and RF is set at 5 to produce probability output. We also adopt four imbalance recovery methods to deal with class imbalance: Random Oversampling (OS), Random Undersampling (US), SMOTE \cite{chawla2002smote}, and Sample Weighting (SW). The first three are sampling methods and the last one is cost-sensitive method, which assigns the weight of the minority class samples as the imbalance ratio and the majority class sample as 1. Because the above methods for imbalance data are independent with the classifier, they can be arbitrarily combined with standard classifiers to deal with class imbalance. We use the simplest imbalance recovery methods for class imbalance problem because our intention is not to select the best imbalance recovery method, but to show that the proposed measured index is generally consistent with the improvement made by the imbalance recovery methods. These methods are implemented by  $imbalanced$-$learn$ toolbox in Python \cite{JMLR:v18:16-365}.

The proposed measures are directly calculated on the whole dataset, such that each minority class sample is associated with an $IBI^3$ value and each dataset is associated with a $BI^3$ value. To show the correlation with the standard classifiers with imbalance recovery methods, we carry out 10-fold cross validation with 5 different random partition runs, on each combination of classifier and the imbalance recovery method. Thus, each minority class sample can be calculated as a test sample in its own fold and averaged by 5 runs. The correlation analysis is conducted in two levels:
\begin{itemize}
	\item Instance level correlation. All the minority class samples in all datasets are accumulated. We calculate the correlation between $IBI^3$ and the increase of prediction score made by the imbalance recovery methods on each classifier by (\ref{recovery_probability}). In this case, $f'$ is the classifier with imbalance recovery methods and $f$ is the standard classifier. Thus, we can evaluate if $IBI^3$ is consistent to the difference made by the imbalance recovery method on minority class samples.
	\item Data level correlation. All the datasets are accumulated. We calculate the $BI^3$ on each dataset and compare it with the improvement of F1 score made by the imbalance recovery methods. Thus, we can evaluate if $BI^3$ can show the impact of imbalance to the dataset in terms of improvement of F1 score.
\end{itemize}
The number of nearest neighbors $k_0$ is set at 5 for all experiments. Because this is the first work to propose a measure describing the impact degree of imbalanced dataset, there is no proper comparison methods on the same purpose. Thus, we compare with three hardness measures $kDN$ and $CL$ proposed in \cite{smith2014instance} and $CM$ proposed in \cite{anwar2014measurement}. They are related to $kNN$ and Naive Bayes classifier but with no consideration about imbalance. $kDN$ measures the percentage of data point $\textbf{x}$'s neighbors that are not in the same class as $\textbf{x}$:
\begin{align}\nonumber
kDN(\textbf{x},y)=\frac{|\{(\textbf{x}',y'):\textbf{x}'\in kNN(\textbf{x}), y'\ne y\}|}{k}
\end{align}
where $kNN(\textbf{x})$ is the set of $k$ nearest neighbors of $\textbf{x}$ and $|\cdot|$ is the size of the set. We also set $k=5$. CL measures the global overlap between classes and the likelihood of a sample belonging to its opposite class:
\begin{align}\nonumber
CL(\textbf{x},y) = 1- \prod_i^dp(\textbf{x}_i,y)
\end{align}
where $d$ is the number of dimensions and $p(\textbf{x}_i,y)$ is the samples's likelihood on $i$th feature to its class $y$. It uses the same assumption as Naive Bayes that the features are independent between each other. The original version of $CL$ in \cite{smith2014instance} is the likelihood of a sample belonging to its own class. However, to be consistent with other methods in this paper that the measurement is positive correlated with the instance hardness, we therefore use one to subtract the original CL. We average the values of $kDN$ and $CL$ on all minority class samples to get the data level index. $CM$ is a data level complexity measure:
\begin{align}\nonumber
CM(\textbf{x},y) &= I\bigg(\frac{|\{(\textbf{x}',y'):\textbf{x}'\in kNN(\textbf{x}), y'= y\}|}{k}\le 0.5\bigg)\\\nonumber
CM(\mathcal{D}) &=\frac{1}{N}\sum_{i=1}^NCM(\textbf{x}_i,y_i)
\end{align}
where $I$ is the indicator function. For the data level correlation analysis, we also compare with IR, because it is usually regarded as an index to measure the difficulty of an imbalanced dataset. In summary, we compare $IBI^3$ with $kDN$ and $CL$ for instance level correlation, and compare $BI^3$ with $kDN$, $CL$, $CM$ and $IR$ for data level correlation.

\begin{figure}[!t]
	\hspace{-.2cm}
	\subfloat{
	\includegraphics[width=.24\textwidth]{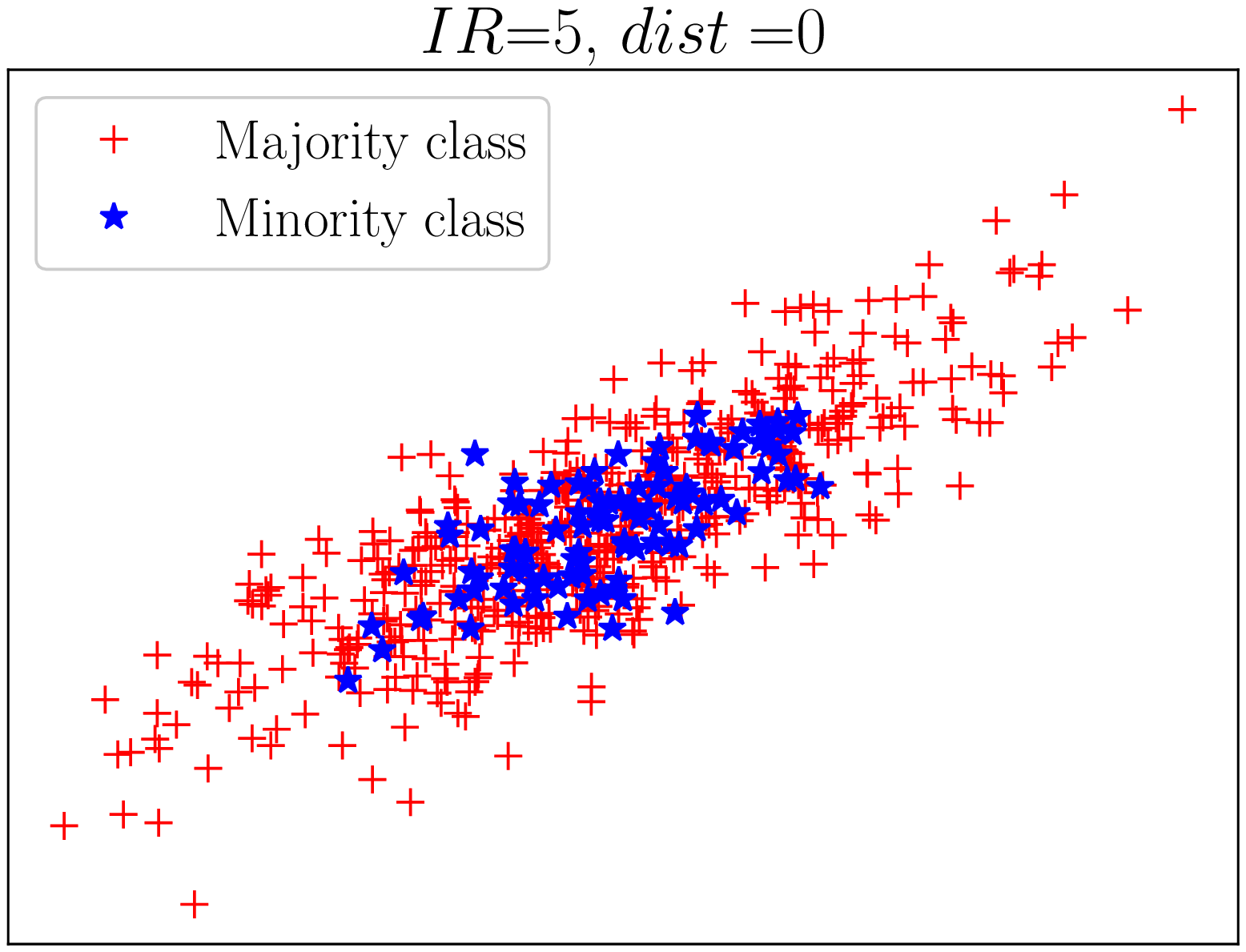}}
	\hspace{-.3cm}
	\subfloat{
	\includegraphics[width=.24\textwidth]{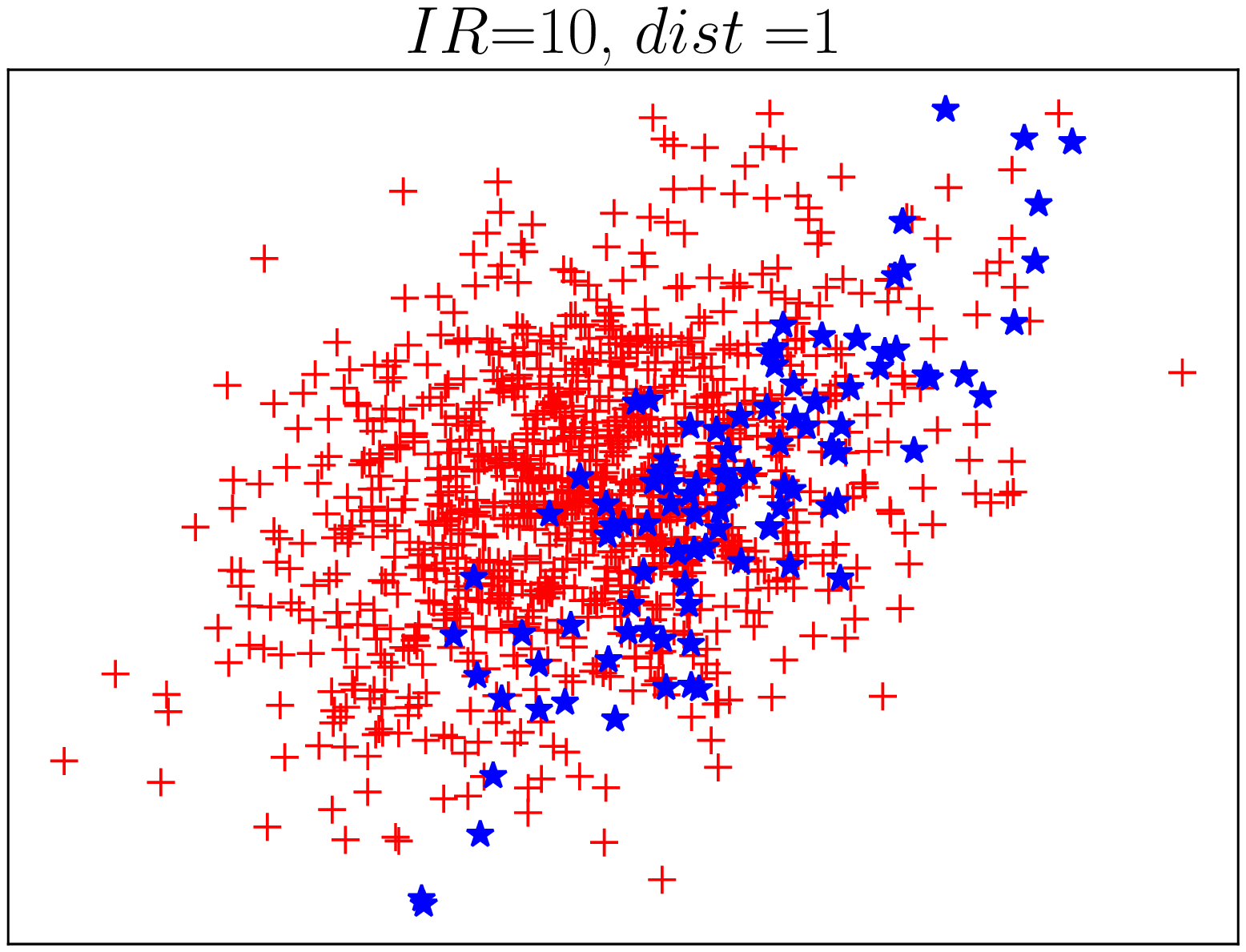}}
	\hspace{-.3cm}
	\subfloat{
	\includegraphics[width=.24\textwidth]{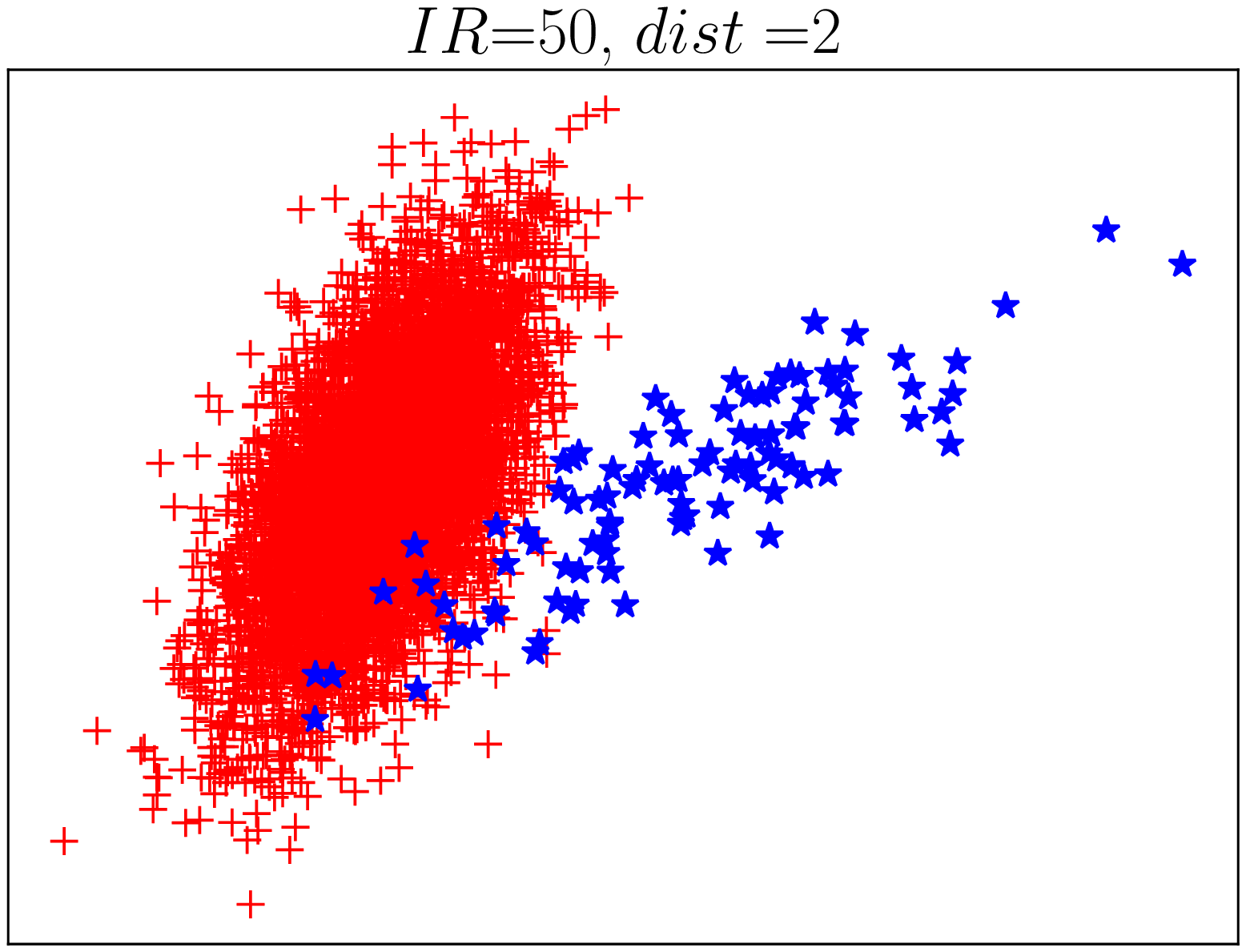}}
	\hspace{-.3cm}
	\subfloat{
	\includegraphics[width=.24\textwidth]{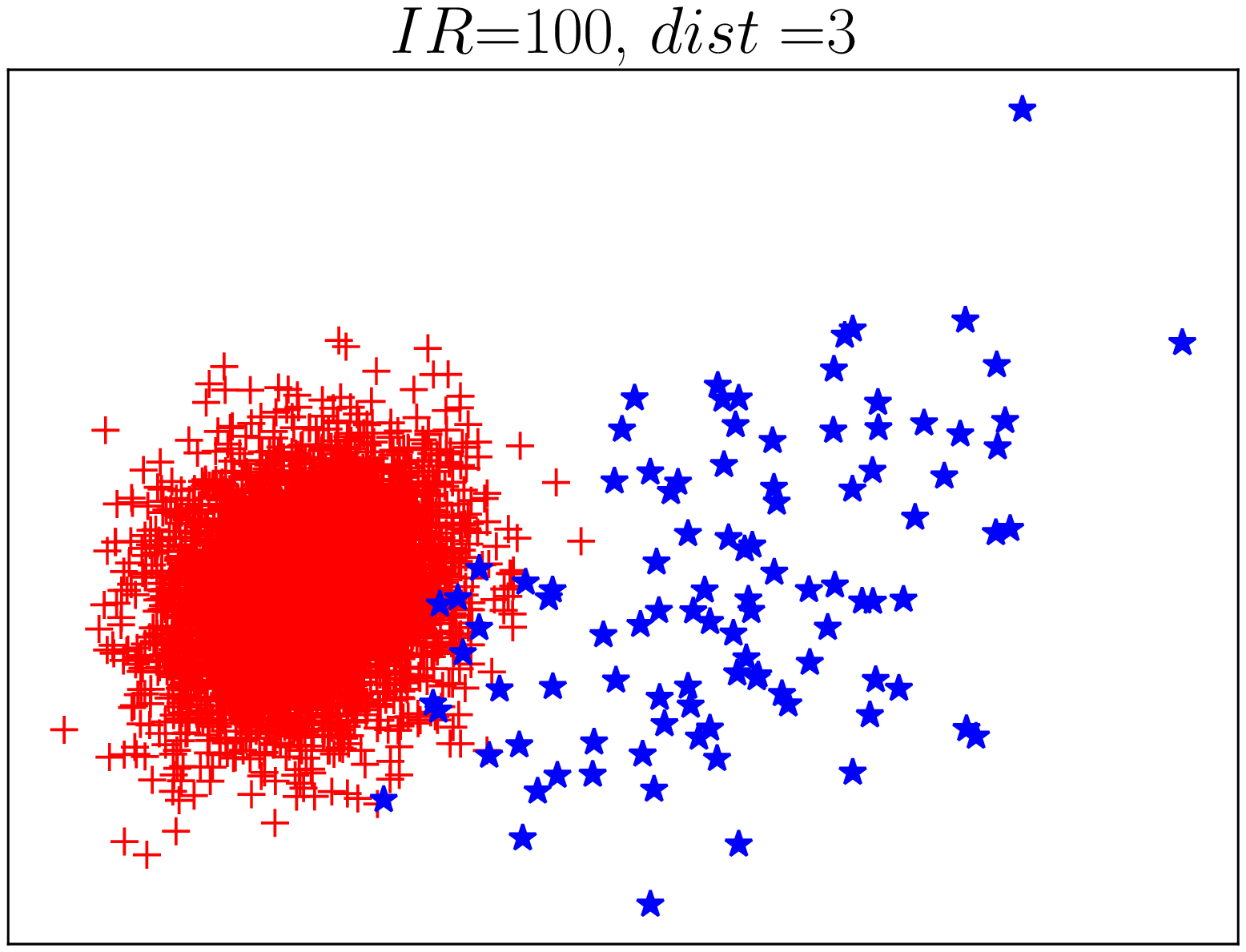}}\\
	\hspace{-.2cm}
	\subfloat{
	\includegraphics[width=.24\textwidth]{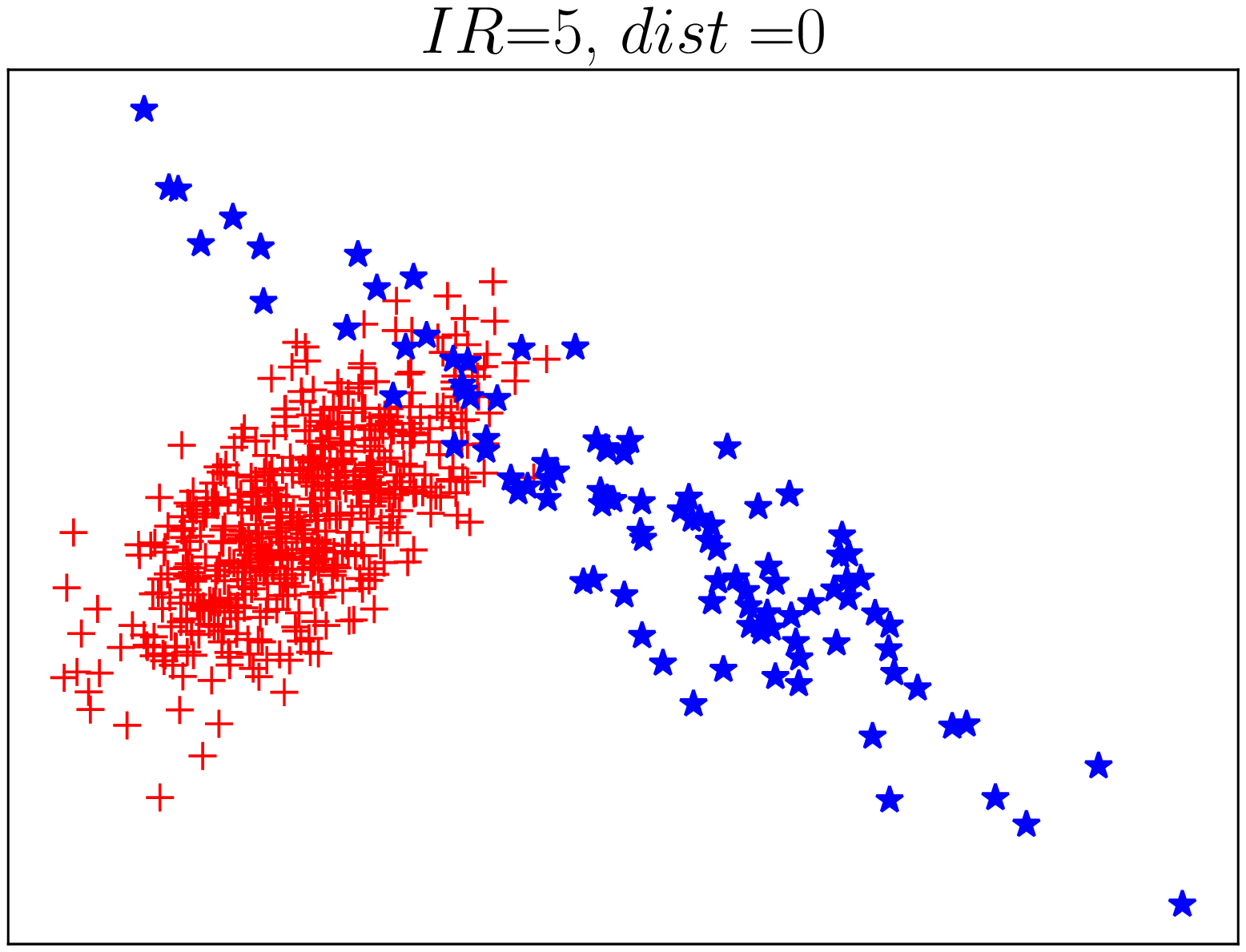}}
	\hspace{-.3cm}
	\subfloat{
	\includegraphics[width=.24\textwidth]{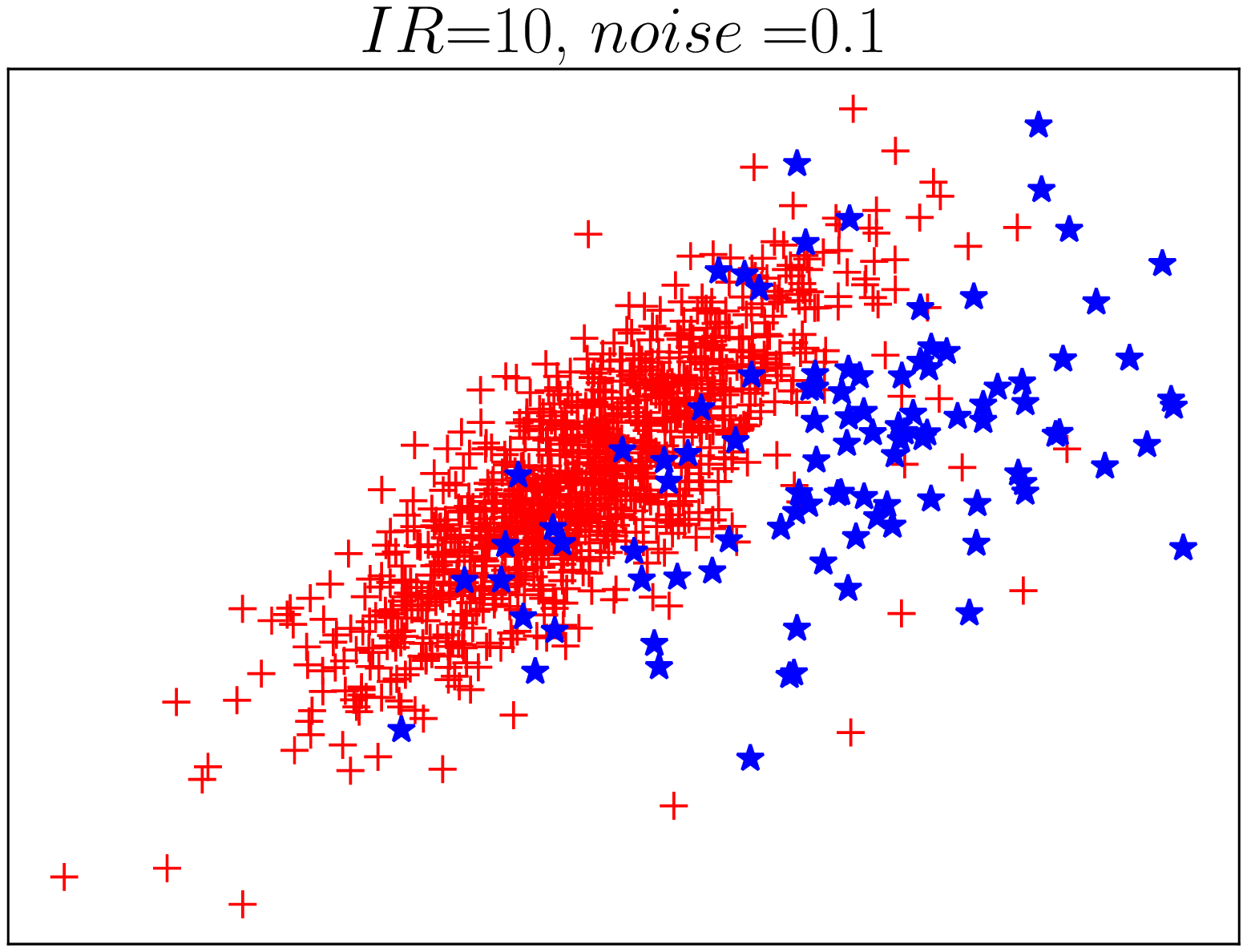}}
	\hspace{-.3cm}
	\subfloat{
	\includegraphics[width=.24\textwidth]{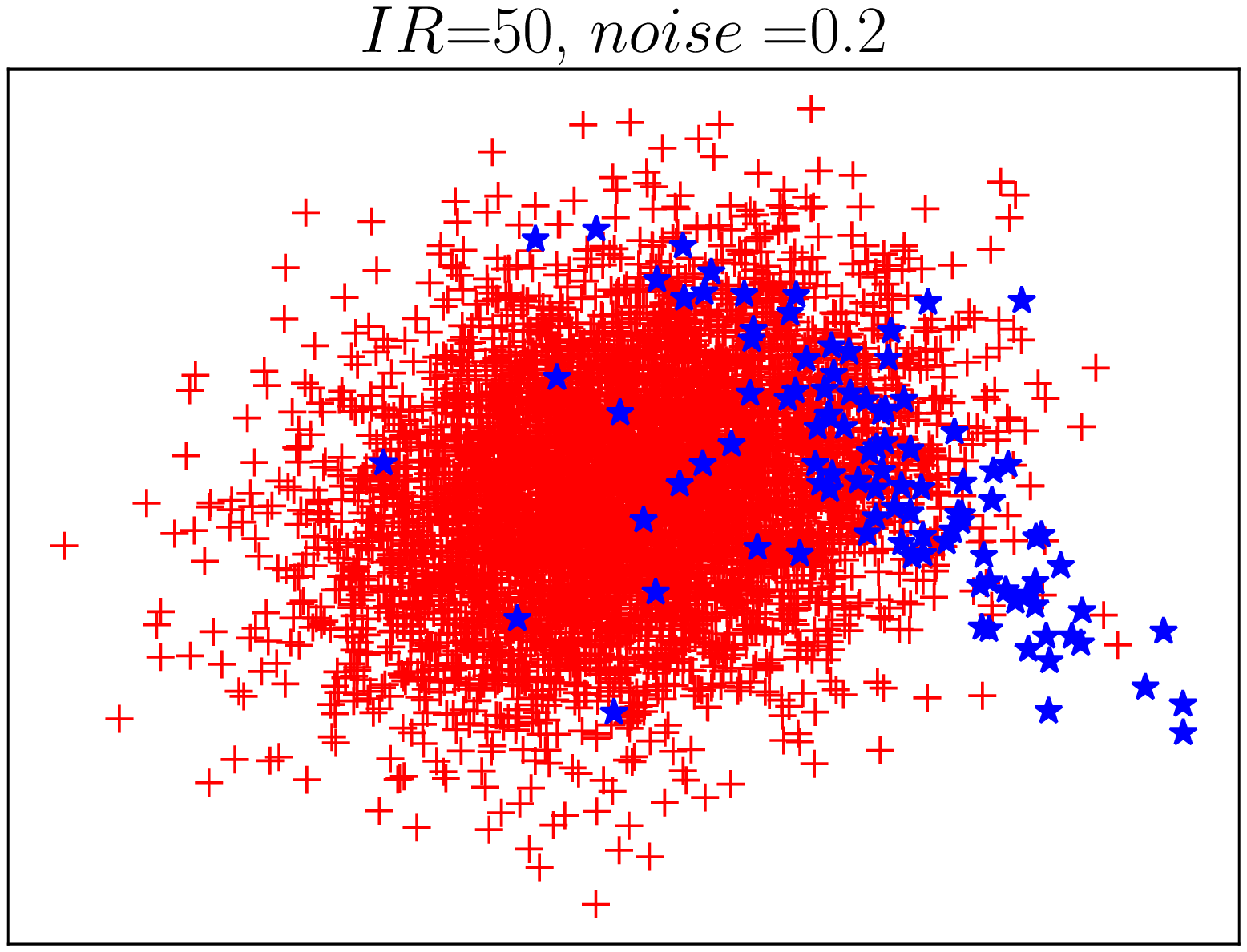}}
	\hspace{-.3cm}
	\subfloat{
	\includegraphics[width=.24\textwidth]{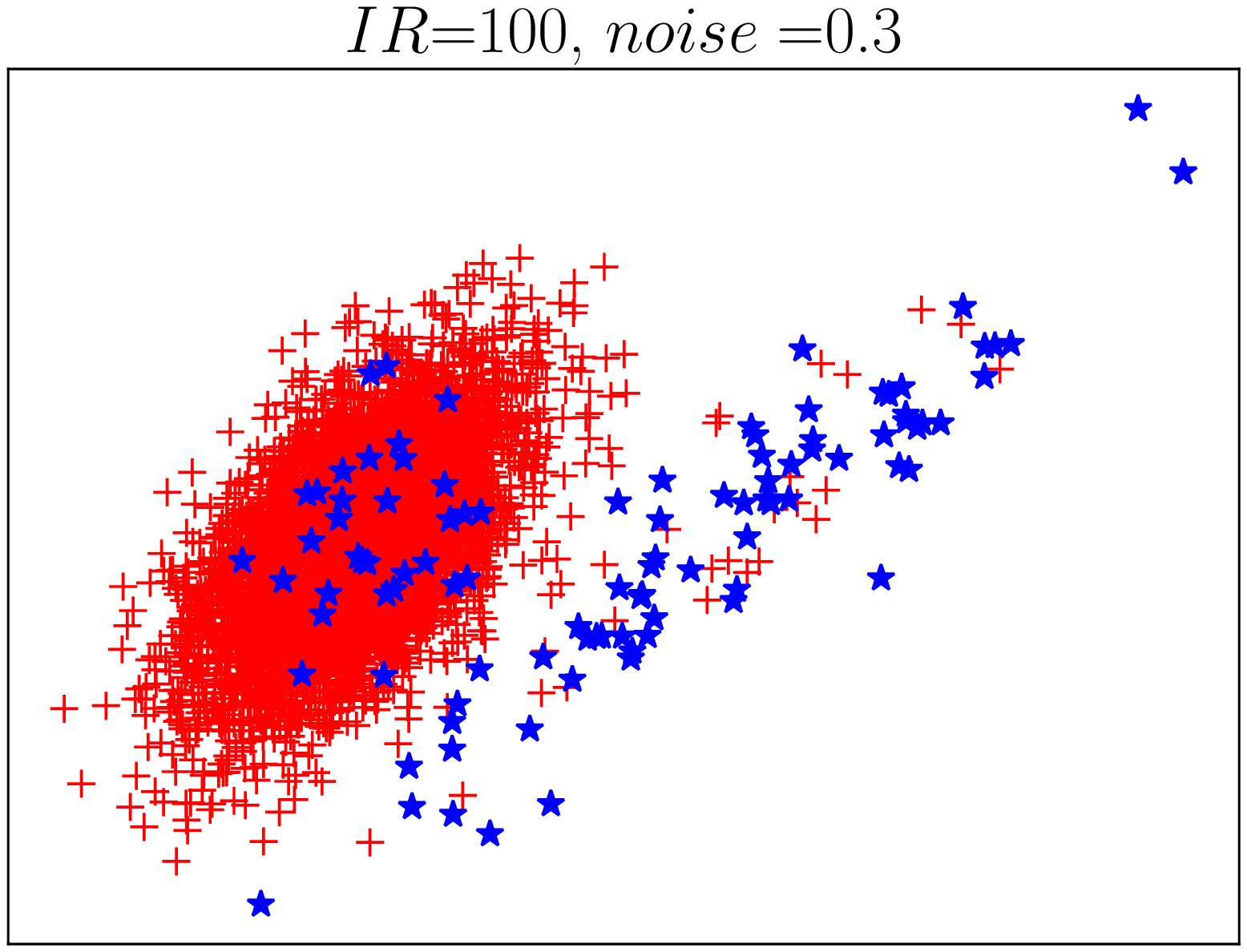}}
	\caption{Eight synthetic binary class imbalanced datasets in dataset group $syn\_overlap$ (upper row) and $syn\_noise$ (lower row) with different covariance combination.}
	\label{syn_overlap}
\end{figure}

\subsection{Synthetic Data}
We first evaluate the proposed index on synthetic binary class datasets. Two group synthetic datasets are generated:
\begin{enumerate}
	\item $syn\_overlap$: The between-class distance and IR are adjusted.
	\item $syn\_noise$: The noise level and IR are adjusted.
\end{enumerate}
Both data sats has two classes that are generated from normal distribution with 2 dimensions. The number of samples in the minority class $N_p$ is fixed at 100 and the number of samples in the majority class $N_n$ varies in the set $\{500, 1000, 5000\}$, where IRs are 5, 10 and 50, respectively. For dataset group $syn\_overlap$, the distance between two classes $dist$ varies in the set $\{0, 1, 2, 3\}$ and there is no noise. For dataset group $syn\_noise$, the noise level $noise$ varies in the set $\{0, 0.1, 0.2, 0.3\}$, where 0.1 means that there are 10\% of the minority class samples are labelled as the majority class and the same number of the majority class samples are labelled as the minority class. The distance between two classes for dataset group $syn\_noise$ is fixed at 2. For both datasets, the covariance matrix for each class is set to
\begin{align}\nonumber
\Sigma = \left[\begin{array}{cc}\sigma_{11} & \sigma_{12} \\\sigma_{21} & \sigma_{22}\end{array}\right] + 0.1I
\end{align}
where $\sigma_{11},\sigma_{22}\in[0,1]$ and $\sigma_{12},\sigma_{21}\in[-1,1]$ are uniformly random number. The extra term $0.1I$ is to ensure that the covariance matrix is positive semidefinite. The covariance matrix for the positive and negative class are set differently, and the covariance matrix is drawn 10 times to produce different combinations. Therefore, totally there are two groups of $3\times 4 \times 10=120$ datasets with different degree of overlapping, different IR, different noise level, and different covariance. Four of the datasets in dataset group $syn\_overlap$ and four of the datasets in dataset group $syn\_noise$ are shown in Figure \ref{syn_overlap}.

\begin{table}[!t]
	\begin{center}
		\begin{tabular}{cccccc}
			    \Xhline{3\arrayrulewidth}
    &&OS&US&SMOTE&SW\\
\hline
\multirow{4}{*}{$kDN$}&SVM    &{ }0.7627    &{ }0.7840    &{ }0.7506    &{ }0.5285\\
&CART    &-0.0061    &{ }0.7379    &{ }0.4182    &{ }0.2091\\
&5NN    &{ }0.2200    &{ }0.8485    &{ }0.5801    &{ }0.2925\\
&RF    &{ }0.0971    &{ }0.7846    &{ }0.4572    &{ }0.3515\\
&AdaBoost    &{ }0.2158    &-0.2363    &{ }0.2187    &{ }0.2156\\
\hline
\multirow{4}{*}{$CL$}&SVM    &{ }0.6016    &{ }0.6031    &{ }0.5939    &{ }0.4431\\
&CART    &-0.0576    &{ }0.5578    &{ }0.3964    &{ }0.2188\\
&5NN    &{ }0.2453    &{ }0.5930    &{ }0.4695    &{ }0.2803\\
&RF    &{ }0.2002    &{ }0.6312    &{ }0.4784    &{ }0.3738\\
&AdaBoost    &{ }0.1314    &-0.2348    &{ }0.1696    &{ }0.1267\\
\hline
\multirow{4}{*}{$IBI^3$}&SVM    &{ }\textbf{0.8501}    &{ }\textbf{0.8512}    &{ }\textbf{0.8416}    &{ }\textbf{0.5977}\\
&CART    &{ }\textbf{0.1105}    &{ }\textbf{0.8072}    &{ }\textbf{0.5881}    &{ }\textbf{0.3522}\\
&5NN    &{ }\textbf{0.4995}    &{ }\textbf{0.9311}    &{ }\textbf{0.7997}    &{ }\textbf{0.5965}\\
&RF    &{ }\textbf{0.3215}    &{ }\textbf{0.8531}    &{ }\textbf{0.6769}    &{ }\textbf{0.5487}\\
&AdaBoost    &{ }\textbf{0.2841}    &\textbf{-0.0944}    &{ }\textbf{0.2664}    &{ }\textbf{0.2815}\\
   \Xhline{3\arrayrulewidth}
		\end{tabular}
	\end{center}
	\caption{The instance level Spearman ranked correlation between the indices and the prediction score increase of minority class sample on datasets group $syn\_overlap$. The highest correlation is shown in bold face.}
	\label{syn_instance_table}
\end{table}%

\subsubsection{Results on dataset group $syn\_overlap$}
The instance level correlation is shown in Table \ref{syn_instance_table}. Generally, $IBI^3$ shows higher correlation than $kDN$ and $CL$. $IBI^3$ shows highest correlations on SVM with OS, US and SMOTE, which are generally more than 0.85. The high correlation means that if the prediction score of a minority class sample can be increased by SVM with the imbalance recovery methods, its $IBI^3$ is also high. Both $IBI^3$ and $kDN$ utilize the nearest neighbors to calculate the measure. $kDN$ has much lower correlation compared with $IBI^3$, because the imbalance factor is not considered in $kDN$. The correlation on CART with OS is not high for all indices, though $IBI^3$ achieves the highest one 0.1105 and other two methods have negative correlations. A possible reason is that the random oversampling simply duplicates the minority class samples so that the leaf node of the decision tree is full of the duplicated the minority class samples after oversampling, which does not increase the prediction score of the minority class samples. Meanwhile, CART with US has high correlation with $IBI^3$, which may suggest that US is the more effective way to increase the minority class prediction score with CART. It can be noticed that on 5NN, the correlations of $IBI^3$ of OS and SW are lower than the ones of US and SMOTE. A possible reason is that OS and SW only work if the training the minority class samples are in the neighborhood of the testing the minority class sample. If the testing the minority class sample are surrounded by training the majority class samples, it will still be misclassified, because OS and SW only duplicate and increase the weight of the training the minority class samples. For RF, the correlation of $IBI^3$ is higher than CART, because the ensemble of trees is more robust to increase the prediction score, especially for US which shows 0.8531 correlation with $IBI^3$. For AdaBoost, the correlation is low for all indices with all imbalance recovery methods. By investigation, we found that the minority class prediction score of AdaBoost is very close to 0.5 and the imbalance recovery methods only increase the score a little to make it over 0.5 which will change the classification result. Therefore, AdaBoost has small correlation with the indices.

\begin{table}[!t]
	\begin{center}
		\begin{tabular}{cccccc}
			    \Xhline{3\arrayrulewidth}
    &&OS&US&SMOTE&SW\\
\hline
\multirow{4}{*}{$kDN$}&SVM    &{ }0.6883    &{ }0.6754    &{ }0.7036    &{ }0.6938\\
&CART    &{ }0.3656    &{ }0.5782    &{ }0.4497    &{ }0.4337\\
&5NN    &{ }0.3216    &{ }0.5628    &{ }0.4454    &{ }0.3985\\
&RF    &{ }0.4863    &{ }0.6647    &{ }0.5672    &{ }0.4918\\
&AdaBoost    &{ }0.5804    &{ }0.5601    &{ }0.5905    &{ }0.5821\\
\hline
\multirow{4}{*}{$CL$}&SVM    &{ }0.6731    &{ }0.6478    &{ }0.6894    &{ }0.6786\\
&CART    &{ }0.4420    &{ }0.5536    &{ }0.4860    &{ }0.4814\\
&5NN    &{ }0.4311    &{ }0.5477    &{ }0.4940    &{ }0.4611\\
&RF    &{ }0.5378    &{ }0.6148    &{ }0.5737    &{ }0.5347\\
&AdaBoost    &{ }0.4346    &{ }0.4156    &{ }0.4260    &{ }0.4388\\
\hline
\multirow{4}{*}{$CM$}&SVM    &{ }0.3600    &{ }0.3346    &{ }0.3753    &{ }0.3655\\
&CART    &{ }0.2650    &{ }0.2357    &{ }0.1693    &{ }0.2184\\
&5NN    &{ }0.2183    &{ }0.2407    &{ }0.1809    &{ }0.1866\\
&RF    &{ }0.3793    &{ }0.3270    &{ }0.2956    &{ }0.3999\\
&AdaBoost    &{ }0.2398    &{ }0.1664    &{ }0.2206    &{ }0.2338\\
\hline
\multirow{4}{*}{$IR$}&SVM    &{ }0.3312    &{ }0.3540    &{ }0.3324    &{ }0.3324\\
&CART    &{ }0.1909    &{ }0.3674    &{ }0.3494    &{ }0.2958\\
&5NN    &{ }0.1811    &{ }0.3671    &{ }0.3203    &{ }0.2849\\
&RF    &{ }0.1538    &{ }0.3459    &{ }0.3061    &{ }0.1461\\
&AdaBoost    &{ }0.3742    &{ }0.4403    &{ }0.4154    &{ }0.3844\\
\hline
\multirow{4}{*}{$BI^3$}&SVM    &{ }\textbf{0.7764}    &{ }\textbf{0.7710}    &{ }\textbf{0.7900}    &{ }\textbf{0.7807}\\
&CART    &{ }\textbf{0.4560}    &{ }\textbf{0.6883}    &{ }\textbf{0.5716}    &{ }\textbf{0.5485}\\
&5NN    &{ }\textbf{0.4263}    &{ }\textbf{0.6757}    &{ }\textbf{0.5682}    &{ }\textbf{0.5219}\\
&RF    &{ }\textbf{0.5682}    &{ }\textbf{0.7587}    &{ }\textbf{0.6709}    &{ }\textbf{0.5682}\\
&AdaBoost    &{ }\textbf{0.6910}    &{ }\textbf{0.6998}    &{ }\textbf{0.7101}    &{ }\textbf{0.6951}\\
   \Xhline{3\arrayrulewidth}
		\end{tabular}
	\end{center}
	\caption{The data level Spearman ranked correlation between the indices and the improvement of F1 score by different imbalance recovery methods on datasets group $syn\_overlap$. The highest correlation is shown in bold face.}
	\label{syn_data_table}
\end{table}%

\begin{table}[!t]
	\begin{center}
		\begin{tabular}{lcccc}
		\Xhline{3\arrayrulewidth}
		&$dist=0$&$dist=1$&$dist=2$&$dist=3$\\
		\hline
			$IR=5$&0.2646  & 0.2037  & 0.1055  & 0.0332\\
			$IR=10$&0.3696  & 0.2895  & 0.1580  & 0.0505\\
			$IR=50$&0.5120  & 0.4639  & 0.2593  & 0.1119\\
		\Xhline{3\arrayrulewidth}
		\end{tabular}
	\end{center}
	\caption{The value of $BI^3$ on dataset group $syn\_overlap$ averaged over 10 different variances.}
	\label{avg_BI}
\end{table}%

The data level correlation is shown in Table \ref{syn_data_table}. $BI^3$ shows the highest correlation with the improvement of F1 score with all classifier and all imbalance recovery methods, where the correlations are generally greater than 0.5. For SVM, $BI^3$ shows high correlations with all imbalance recovery methods. All the correlations are greater than 0.77. CART, 5NN and RF also show high correlation compared with other indices. It is interesting to notice that AdaBoost has the generally second high correlation over all imbalance recovery methods. However, its instance level correlation is very low as shown in Table \ref{syn_instance_table}. As explained before, the increase of prediction score of AdaBoost is little but it changes the prediction and thus influences the F1 score. The correlations of $kDN$ and $CL$ are generally 0.1 less than the ones of $BI^3$, because they do not consider the imbalance into the index. They use pure data complexity to describe the effect caused by imbalance, and are thus not as accurate as $BI^3$. $CM$ shows low correlations because it sums up the neighborhood indicator values of all the majority and minority class samples. It can be used to represent the overall classification complexity of a dataset, but cannot show the impact of imbalance to it. For data level correlation, $IR$ is also compared as an index. However, most correlations between $IR$ and the imbalance recovery methods are lower than 0.4. That means $IR$ can be hardly used as an index to describe the influence of class imbalance problem.

In summary, on dataset group $syn\_overlap$, $BI^3$ shows high correlation with the improvement of F1 score by imbalance recovery methods on all classifiers. It means that the value of $BI^3$ is a proper index to describe how much improvement of F1 score can be made by applying imbalance recovery methods. In other words, if a dataset has low $BI^3$ value, it should be carefully considered whether or not to apply imbalance recovery methods because the improvement is limited or even negative.  Table \ref{avg_BI} shows the value of $BI^3$ averaged over 10 different variances on dataset group $syn\_overlap$. It can be observed that as the distance between two classes increases, $BI^3$ decreases because the overlapping region is reduced. In addition, when $IR$ is increasing, $BI^3$ is also increased. When $dist=3$ and $IR=50$, where the two classes are seldom overlapped, the value of $BI^3$ is comparable with $dist=2$ and $IR=5$. Therefore, it verifies again that $IR$ is not the only cause to make classification performance degenerated and $BI^3$ is more proper to describe the impact brought by imbalance.

\begin{table}[!t]
\begin{center}
\begin{tabular}{cccccc}
        \Xhline{3\arrayrulewidth}
    &&OS&US&SMOTE&SW\\
\hline
\multirow{4}{*}{$kDN$}&SVM    &{ }0.5958    &{ }0.6488    &{ }0.5856    &{ }0.3945\\
&CART    &-0.0517    &{ }0.5487    &{ }0.2505    &{ }0.1050\\
&5NN    &{ }0.1565    &{ }0.7114    &{ }0.4406    &{ }0.2298\\
&RF    &-0.0442    &{ }0.6193    &{ }0.2335    &{ }0.1269\\
&AdaBoost    &{ }0.1323    &-0.4109    &{ }0.1510    &{ }0.1195\\
\hline
\multirow{4}{*}{$CL$}&SVM    &{ }0.4814    &{ }0.5104    &{ }0.4749    &{ }0.4822\\
&CART    &{ }0.1185    &{ }0.3116    &{ }0.1503    &{ }0.0186\\
&5NN    &{ }0.0068    &{ }0.3447    &{ }0.2026    &{ }0.0245\\
&RF    &{ }0.0587    &{ }0.4125    &{ }0.1903    &{ }0.0281\\
&AdaBoost    &{ }0.0039    &-0.4974    &{ }0.0371    &{ }0.0266\\
\hline
\multirow{4}{*}{$IBI^3$}&SVM    &{ }\textbf{0.7283}    &{ }\textbf{0.7421}    &{ }\textbf{0.7222}    &{ }\textbf{0.4516}\\
&CART    &{ }\textbf{0.1836}    &{ }\textbf{0.6984}    &{ }\textbf{0.4868}    &{ }\textbf{0.3605}\\
&5NN    &{ }\textbf{0.5170}    &{ }\textbf{0.9150}    &{ }\textbf{0.7487}    &{ }\textbf{0.6372}\\
&RF    &{ }\textbf{0.3223}    &{ }\textbf{0.7763}    &{ }\textbf{0.5727}    &{ }\textbf{0.4784}\\
&AdaBoost    &{ }\textbf{0.2358}    &\textbf{-0.1407}    &{ }\textbf{0.1957}    &{ }\textbf{0.2255}\\
   \Xhline{3\arrayrulewidth}
\end{tabular}
\end{center}
\caption{The instance level Spearman ranked correlation between the indices and the prediction score increase of minority class sample on datasets group $syn\_noise$. The highest correlation is shown in bold face.}
\label{syn_noise_instance_table}
\end{table}

\begin{table}[!t]
\begin{center}
\begin{tabular}{cccccc}
    \Xhline{3\arrayrulewidth}
    &&OS&US&SMOTE&SW\\
\hline
\multirow{4}{*}{$kDN$}&SVM    &{ }0.6785    &{ }0.6748    &{ }0.6750    &{ }0.6888\\
&CART    &{ }0.4744    &{ }0.3890    &{ }0.3046    &{ }0.4541\\
&5NN    &{ }0.4755    &{ }0.5358    &{ }0.4290    &{ }0.4196\\
&RF    &{ }0.6739    &{ }0.6245    &{ }0.5762    &{ }0.6911\\
&AdaBoost    &{ }0.6793    &{ }0.4907    &{ }0.6521    &{ }0.6811\\
\hline
\multirow{4}{*}{$CL$}&SVM    &{ }0.4504    &{ }0.4382    &{ }0.4459    &{ }0.4598\\
&CART    &{ }0.1943    &{ }0.0798    &{ }0.0039    &{ }0.1455\\
&5NN    &{ }0.2151    &{ }0.2783    &{ }0.1707    &{ }0.1072\\
&RF    &{ }0.4557    &{ }0.3545    &{ }0.3325    &{ }0.4797\\
&AdaBoost    &{ }0.4062    &{ }0.1945    &{ }0.3839    &{ }0.4051\\
\hline
\multirow{4}{*}{$CM$}&SVM    &-0.0050    &-0.0214    &{ }0.0019    &{ }0.0001\\
&CART    &-0.2560    &-0.2024    &-0.3832    &-0.3139\\
&5NN    &-0.2430    &-0.1333    &-0.2233    &-0.3631\\
&RF    &{ }0.0313    &-0.0439    &-0.0812    &{ }0.0628\\
&AdaBoost    &-0.0750    &-0.2031    &-0.0503    &-0.0795\\
\hline
\multirow{4}{*}{$IR$}&SVM    &{ }0.4561    &{ }0.4750    &{ }0.4496    &{ }0.4567\\
&CART    &{ }0.6240    &{ }0.4997    &{ }0.5161    &{ }0.6495\\
&5NN    &{ }0.5575    &{ }0.5094    &{ }0.5059    &{ }0.6491\\
&RF    &{ }0.4237    &{ }0.4688    &{ }0.4770    &{ }0.3975\\
&AdaBoost    &{ }0.5265    &{ }0.5463    &{ }0.4897    &{ }0.5358\\
\hline
\multirow{4}{*}{$BI^3$}&SVM    &{ }\textbf{0.7781}    &{ }\textbf{0.7806}    &{ }\textbf{0.7729}    &{ }\textbf{0.7865}\\
&CART    &{ }\textbf{0.6661}    &{ }\textbf{0.5588}    &{ }\textbf{0.6168}    &{ }\textbf{0.6613}\\
&5NN    &{ }\textbf{0.6689}    &{ }\textbf{0.6725}    &{ }\textbf{0.6033}    &{ }\textbf{0.6503}\\
&RF    &{ }\textbf{0.7733}    &{ }\textbf{0.7478}    &{ }\textbf{0.7114}    &{ }\textbf{0.7781}\\
&AdaBoost    &{ }\textbf{0.8045}    &{ }\textbf{0.6571}    &{ }\textbf{0.7720}    &{ }\textbf{0.8104}\\
   \Xhline{3\arrayrulewidth}
\end{tabular}
\end{center}
\caption{The data level Spearman ranked correlation between the indices and the improvement of F1 score by different imbalance recovery methods on datasets group $syn\_noise$. The highest correlation is shown in bold face.}
\label{syn_noise_data_table}
\end{table}

\begin{table}[!ht]
	\begin{center}
		\begin{tabular}{lcccc}
		\Xhline{3\arrayrulewidth}
		&$noise=0$&$noise=0.1$&$noise=0.2$&$noise=0.3$\\
		\hline
			$IR=5$&0.0803  & 0.1487  & 0.1988  & 0.2429\\
			$IR=10$&0.1156  & 0.1927  & 0.2529  & 0.3061\\
			$IR=50$&0.2261  & 0.2929  & 0.3446  & 0.3978\\
		\Xhline{3\arrayrulewidth}
		\end{tabular}
	\end{center}
	\caption{The value of $BI^3$ on dataset group $syn\_noise$ averaged over 10 different variances.}
	\label{avg_noise_BI}
\end{table}%

\subsubsection{Results on dataset group $syn\_noise$}
The instance level correlation is shown in Table \ref{syn_noise_instance_table}. Same as the results on $syn\_overlap$ $IBI^3$ also shows the highest correlations. However, it can be noticed that the correlations of SVM, CART, RF and AdaBoost are generally lower than the ones of $syn\_overlap$ shown in Table \ref{syn_instance_table}. However, the correlations of 5NN of $syn\_noise$ is comparable with the ones of $syn\_overlap$. The reason is that $IBI^3$ is based on $kNN$ and some minority class noises in the deep region of the majority class receives low $IBI^3$ value according to (\ref{recovery_probability}). However, the prediction score of classifiers like SVM and RF on these noised points will be significantly different if imbalance recovery methods are applied. Therefore, it makes the correlations lower than the ones of $syn\_overlap$. Similarly, $kDN$ also has lower correlations compared with the ones of $syn\_overlap$. The correlations of $CL$ is low because it is based on naive bayes. When there are noises in the dataset, the mean and variance cannot be well estimated and therefore the correlations are also low.

The data level correlation is shown in Table \ref{syn_noise_data_table}. Most of the correlations of $BI^3$ are greater than 0.6. $CL$ has very low correlations with the improvement of F1 score because it is sensitive to the noises. $CM$ even generates negative correlations, which means it is not a proper index to describe the imbalance extent of a noised dataset. Surprisedly, $IR$ shows comparable correlations with $kDN$. It means that if the factor of overlapping is fixed, $IR$ can still partially represent the impact of imbalance to the dataset, although there exists noises.

Table \ref{avg_noise_BI} shows the value of $BI^3$ averaged over 10 different variances on dataset group $syn\_noise$. It can be observed that as the noise level increases or $IR$ increases, the index value also increases. It can be observed that both $IR$ and the noise level play roles on $BI^3$ and thus it verifies again that the performance of classifier on imbalanced dataset depends not only on $IR$.

\setlength{\tabcolsep}{4pt}
\begin{table*}[!t]
\centering
		\begin{tabular}{lcccc|lcccc}
			\Xhline{3\arrayrulewidth}
			dataset&\#Inst.&\#Attr.&$IR$&$BI^3$&dataset&\#Inst.&\#Attr.&$IR$&$BI^3$\\
			\hline
ecoli-0\_vs\_1&220&7&1.86&0.01&                         yeast-1\_vs\_7&459&7&14.30&0.48\\
pima&768&8&1.87&0.10&                                   glass4&214&9&15.46&0.37\\
iris0&150&4&2.00&0.00&                                  ecoli4&336&7&15.80&0.19\\
glass0&214&9&2.06&0.09&                                 abalone9-18&731&8&16.40&0.46\\
yeast1&1484&8&2.46&0.16&                                dermatology-6&358&34&16.90&0.04\\
haberman&306&3&2.78&0.20&                               yeast-1-4-5-8\_vs\_7&693&8&22.10&0.55\\
vehicle2&846&18&2.88&0.10&                              yeast-2\_vs\_8&482&8&23.10&0.24\\
vehicle1&846&18&2.90&0.20&                              flare-F&1066&11&23.79&0.56\\
glass-0-1-2-3\_vs\_4-5-6&214&9&3.20&0.10&               car-good&1728&6&24.04&0.48\\
vehicle0&846&18&3.25&0.09&                              car-vgood&1728&6&25.58&0.37\\
ecoli1&336&7&3.36&0.14&                                 kr-vs-k-one\_vs\_draw&2901&6&26.63&0.12\\
ecoli2&336&7&5.46&0.10&                                 kr-vs-k-one\_vs\_fifteen&2244&6&27.77&0.01\\
segment0&2308&19&6.02&0.02&                             yeast4&1484&8&28.10&0.56\\
glass6&214&9&6.38&0.08&                                 winequality-red-4&1599&11&29.17&0.49\\
yeast3&1484&8&8.10&0.22&                                poker-9\_vs\_7&244&10&29.50&0.47\\
ecoli3&336&7&8.60&0.30&                                 kddcup-guess\_passwd\_vs\_satan&1642&41&29.98&0.00\\
page-blocks0&5472&10&8.79&0.17&                         yeast-1-2-8-9\_vs\_7&947&8&30.57&0.55\\
ecoli-0-3-4\_vs\_5&200&7&9.00&0.11&                     winequality-white-9\_vs\_4&168&11&32.60&0.60\\
yeast-2\_vs\_4&514&8&9.08&0.22&                         yeast5&1484&8&32.73&0.35\\
ecoli-0-6-7\_vs\_3-5&222&7&9.09&0.24&                   kr-vs-k-three\_vs\_eleven&2935&6&35.23&0.08\\
ecoli-0-2-3-4\_vs\_5&202&7&9.10&0.11&                   winequality-red-8\_vs\_6&656&11&35.44&0.48\\
glass-0-1-5\_vs\_2&172&9&9.12&0.43&                     abalone-17\_vs\_7-8-9-10&2338&8&39.31&0.62\\
yeast-0-3-5-9\_vs\_7-8&506&8&9.12&0.34&                 abalone-21\_vs\_8&581&8&40.50&0.50\\
yeast-0-2-5-6\_vs\_3-7-8-9&1004&8&9.14&0.26&            yeast6&1484&8&41.40&0.39\\
yeast-0-2-5-7-9\_vs\_3-6-8&1004&8&9.14&0.14&            winequality-white-3\_vs\_7&900&11&44.00&0.53\\
ecoli-0-4-6\_vs\_5&203&6&9.15&0.11&                     winequality-red-8\_vs\_6-7&855&11&46.50&0.50\\
ecoli-0-1\_vs\_2-3-5&244&7&9.17&0.15&                   kddcup-land\_vs\_portsweep&1061&41&49.52&0.00\\
ecoli-0-2-6-7\_vs\_3-5&224&7&9.18&0.24&                 abalone-19\_vs\_10-11-12-13&1622&8&49.69&0.60\\
ecoli-0-3-4-6\_vs\_5&205&7&9.25&0.11&                   kr-vs-k-zero\_vs\_eight&1460&6&53.07&0.23\\
vowel0&988&13&9.98&0.03&                                winequality-white-3-9\_vs\_5&1482&11&58.28&0.51\\
ecoli-0-6-7\_vs\_5&220&6&10.00&0.21&                    poker-8-9\_vs\_6&1485&10&58.40&0.59\\
glass-0-1-6\_vs\_2&192&9&10.29&0.45&                    shuttle-2\_vs\_5&3316&9&66.67&0.02\\
ecoli-0-1-4-7\_vs\_2-3-5-6&336&7&10.59&0.21&            winequality-red-3\_vs\_5&691&11&68.10&0.60\\
led7digit-0-2-4-5-6-7-8-9\_vs\_1&443&7&10.97&0.20&      abalone-20\_vs\_8-9-10&1916&8&72.69&0.64\\
ecoli-0-1\_vs\_5&240&6&11.00&0.11&                      kddcup-buffer\_overflow\_vs\_back&2233&41&73.43&0.04\\
glass-0-1-4-6\_vs\_2&205&9&11.06&0.47&                  kddcup-land\_vs\_satan&1610&41&75.67&0.02\\
glass2&214&9&11.59&0.46&                                kr-vs-k-zero\_vs\_fifteen&2193&6&80.22&0.07\\
cleveland-0\_vs\_4&173&13&12.31&0.49&                   poker-8-9\_vs\_5&2075&10&82.00&0.72\\
ecoli-0-1-4-6\_vs\_5&280&6&13.00&0.11&                  poker-8\_vs\_6&1477&10&85.88&0.61\\
shuttle-c0-vs-c4&1829&9&13.87&0.01&                     abalone19&4174&8&129.44&0.68\\
\Xhline{3\arrayrulewidth}
		\end{tabular}
\caption{Information of 80 Imbalanced datasets}
\label{real_data}
\end{table*}

\subsection{Real Benchmark Data}

We use 80 real datasets from KEEL dataset repository \cite{alcala2011keel}. The details of the datasets is shown in Table \ref{real_data}. $IR$ ranges from 1.86 to 129.44 over all 80 datasets. For real benchmark data, we also compare the proposed $IBI^3$ and $BI^3$ with $kDN$, $CL$, $CM$ and $IR$, in instance level and data level, respectively.

\begin{figure}[!t]
	\subfloat[]{
	\includegraphics[width=.24\textwidth]{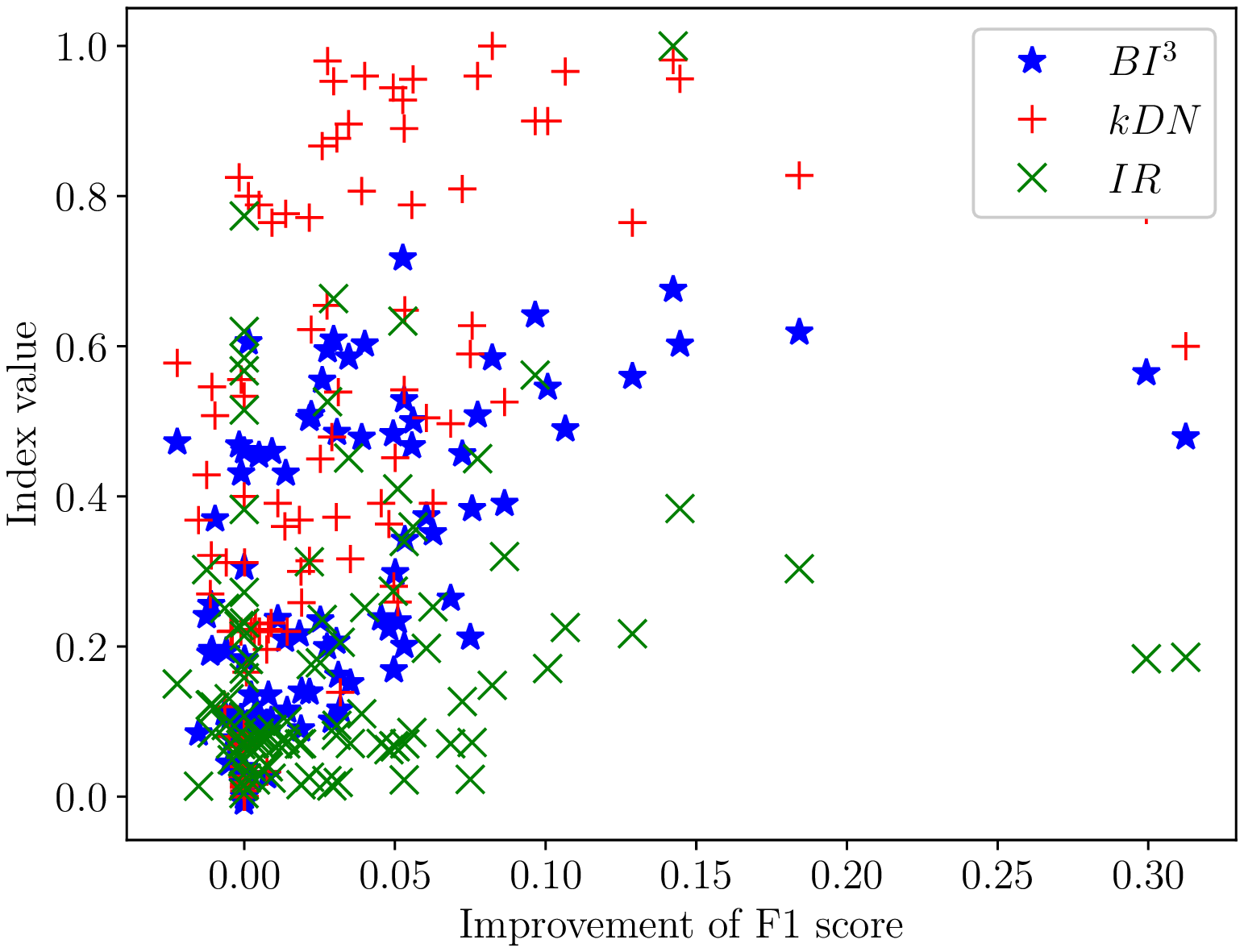}}
	\hspace{-.3cm}
	\subfloat[]{
	\includegraphics[width=.24\textwidth]{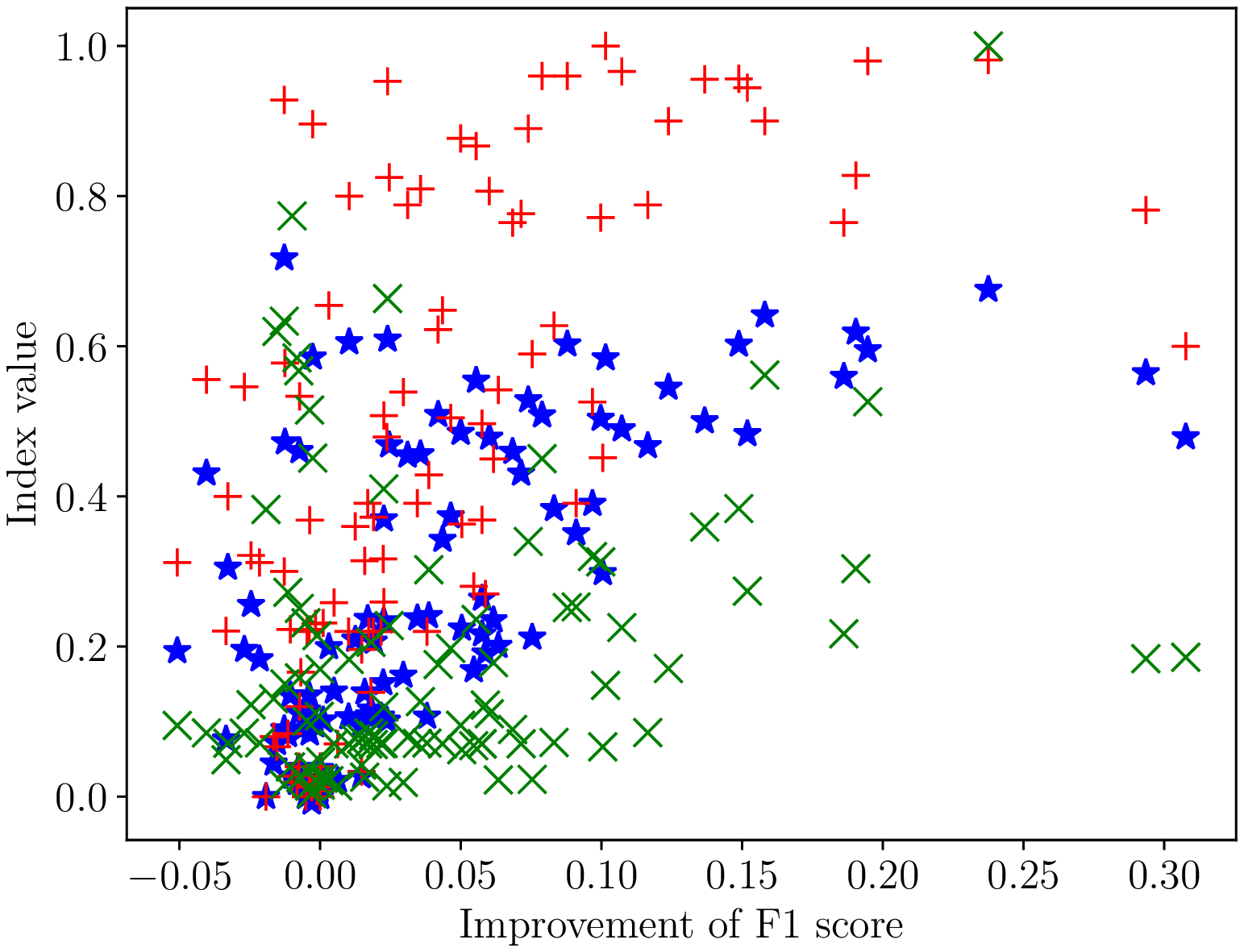}}\\
	\subfloat[]{
	\includegraphics[width=.24\textwidth]{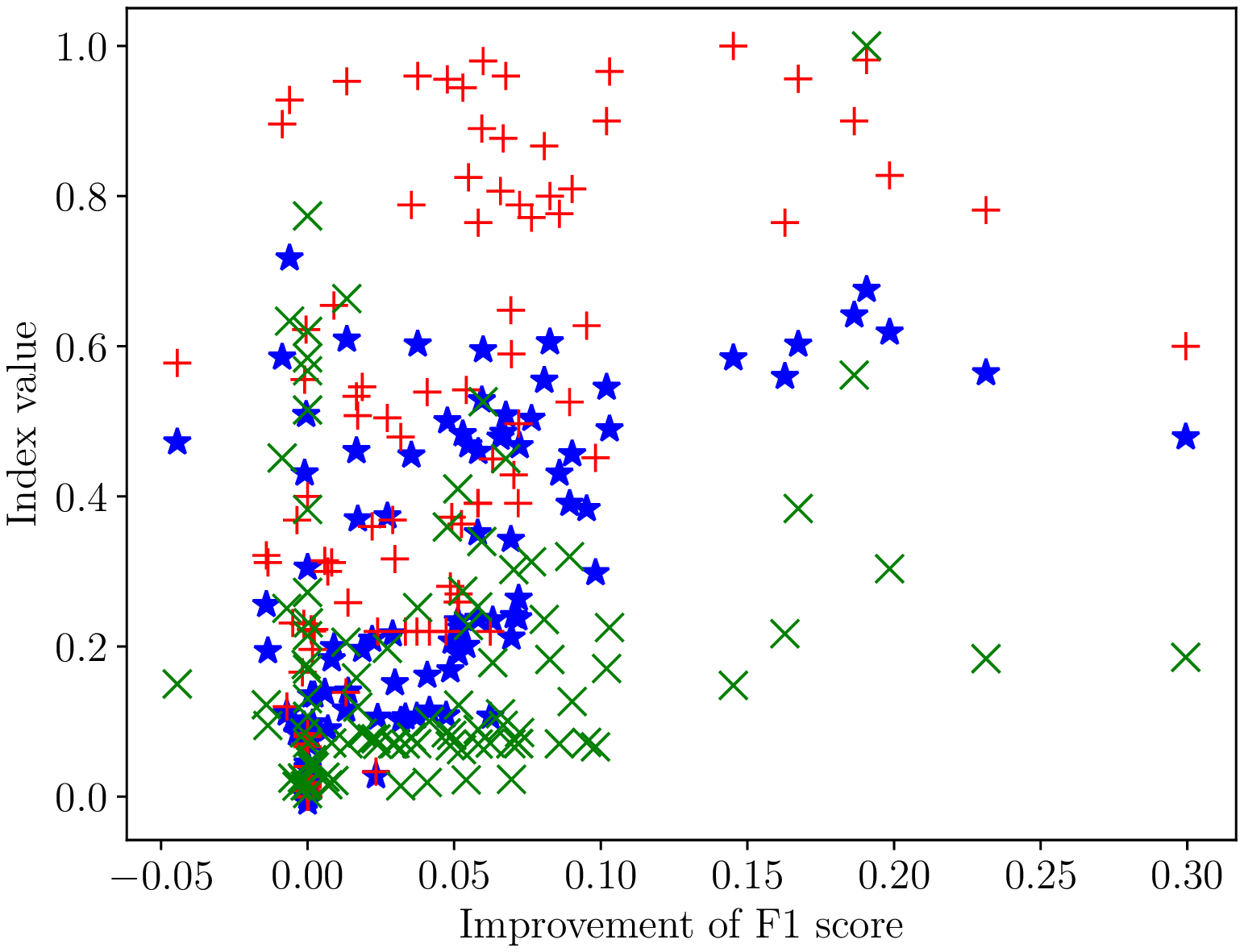}}
	\hspace{-.3cm}
	\subfloat[]{
	\includegraphics[width=.24\textwidth]{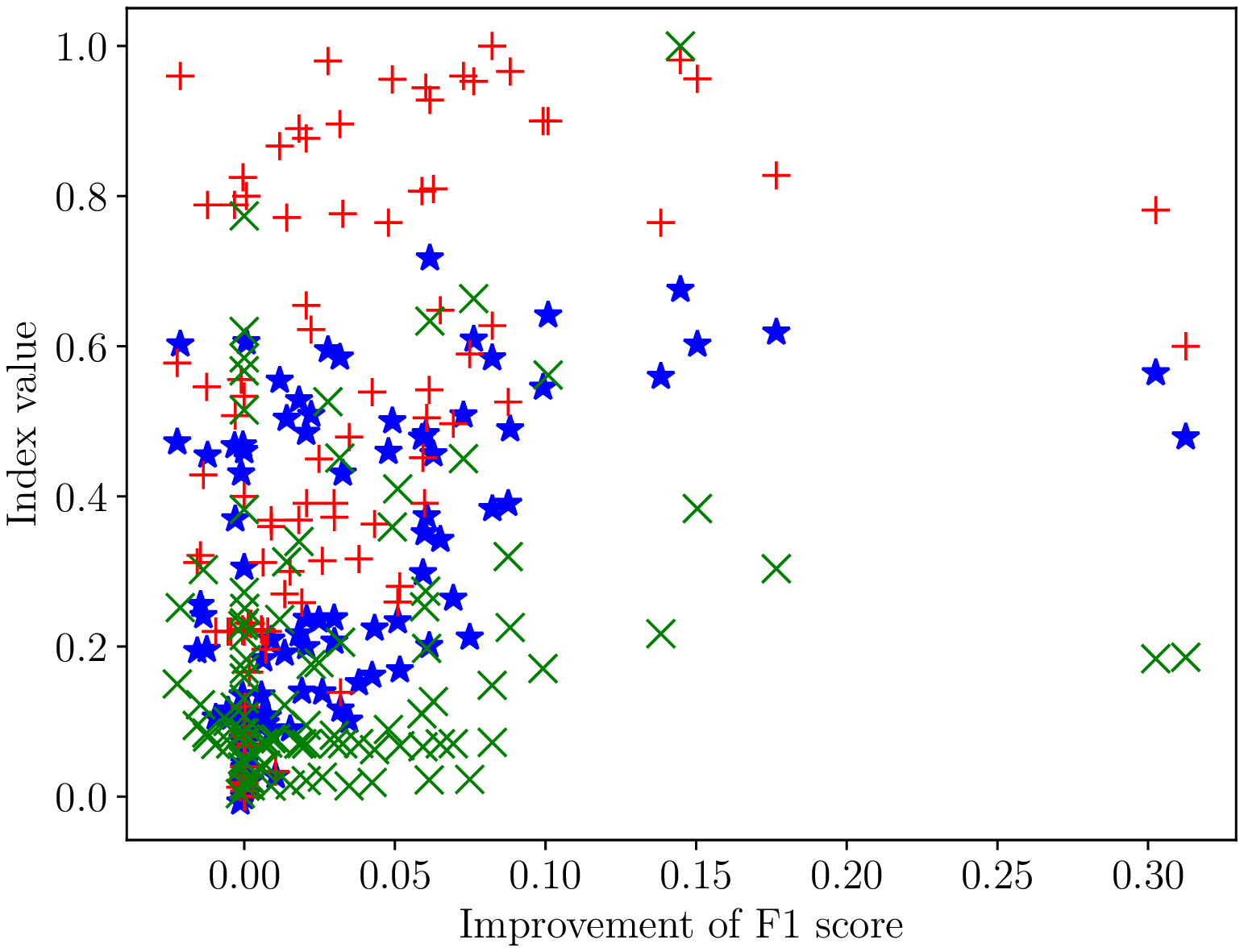}}
	\caption{Index value of $BI^3$, $kDN$, $IR$ over 80 KEEL real benchmark imbalanced datasets on sorted along the percentage of recovered the minority class samples of AdaBoost classifier with (a) OS, (b) US, (c) SMOTE, and (d) SW.}
	\label{real_correlation}
\end{figure}

The instance level correlation is shown in Table \ref{real_instance_table}. $IBI^3$ shows higher correlations than $kDN$ and $CL$, because it considers the imbalance factor into the index. 5NN achieves the highest correlation on all imbalance recovery methods, because $BI^3$ is based on $k$NN, and RF achieves the second highest correlation. On the dimension of imbalance recovery methods, US achieves the highest correlation, where the correlations are greater than 0.5 except with AdaBoost.

The data level correlation is shown in Table \ref{real_data_table}. $BI^3$ achieves the highest correlation and most of the correlations are greater than 0.5, which indicates strong correlation. In other words, given a real dataset, we can calculate $BI^3$ without training and testing to estimate the extend of improvement by using imbalance recovery methods. $kDN$ shows higher correlation than $IR$ in general, which means that the data complexity using nearest neighbor can still better represent the imbalance impact on imbalanced data than referring to imbalance ratio. $CM$ achieves low correlation, which means that $CM$ may be a good data complex measurement for imbalanced data, but not a proper index to describe the imbalance impact. 5NN achieves high correlation on instance level but low correlation on data level. A possible reason is that the imbalance recovery methods applied on 5NN only simply changes the prediction score, but does not effectively improve the F1 score. As same as the situation in synthetic data, AdaBoost shows low correlation on instance level but high correlation on data level. The averaged correlation of AdaBoost over all imbalance recovery methods is higher than other classifiers. It means that $BI^3$ can properly reflect the extend of improvement of F1 score of applying imbalance recovery methods on AdaBoost.

Figure \ref{real_correlation} shows the index value of $BI^3$, $kDN$ and $IR$ over 80 real benchmark datasets on AdaBoost classifier with different imbalance recovery methods. $IR$ is normalized to [0,1] to fit in the figure. It can be observed that the majority of the $IR$ points locates on the bottom, which means that the same level of $IR$ leads to different levels of improvement of F1 score. On the contrary, most of $kDN$ points scatter on the top, which means that $kDN$ tend to overestimate the improvement of F1 score, because it only counts the number of neighbors with different class label for the minority class samples. In comparison, $BI^3$ generally increase as the improvement of F1 score increases as shown in the figure. There are only a few points lie on the region that the improvement of F1 score is close to 0 but $BI^3$ has high values. The reason is that the selected imbalance recovery methods are the simplest ones in the literature which may not be effective to improve the F1 score for all the datasets.

\begin{table}[!t]
	\begin{center}
		\begin{tabular}{cccccc}
			    \Xhline{3\arrayrulewidth}
          &&OS&US&SMOTE&SW\\
    \hline
\multirow{4}{*}{$kDN$}&SVM    &{ }0.3117    &{ }0.5224    &{ }0.3157    &{ }0.1459\\
&CART    &{ }0.0996    &{ }0.5103    &{ }0.1941    &{ }0.2120\\
&5NN    &{ }0.3951    &{ }0.8252    &{ }0.5799    &{ }0.4894\\
&RF    &{ }0.3080    &{ }0.6825    &{ }0.3898    &{ }0.3707\\
&AdaBoost    &{ }0.1963    &-0.0735    &{ }0.2248    &{ }0.1711\\
\hline
\multirow{4}{*}{$CL$}&SVM    &{ }0.1689    &{ }0.3802    &{ }0.2002    &{ }0.0684\\
&CART    &{ }0.1077    &{ }0.3216    &{ }0.1562    &{ }0.1768\\
&5NN    &{ }0.2889    &{ }0.4326    &{ }0.3484    &{ }0.3130\\
&RF    &{ }0.2610    &{ }0.4552    &{ }0.2931    &{ }0.3039\\
&AdaBoost    &{ }0.1336    &{ }0.1391    &{ }0.1842    &{ }0.1367\\
\hline
\multirow{4}{*}{$IBI^3$}&SVM    &{ }\textbf{0.3864}    &{ }\textbf{0.5565}    &{ }\textbf{0.4012}    &{ }\textbf{0.1481}\\
&CART    &{ }\textbf{0.1633}    &{ }\textbf{0.5175}    &{ }\textbf{0.2315}    &{ }\textbf{0.2703}\\
&5NN    &{ }\textbf{0.6018}    &{ }\textbf{0.8981}    &{ }\textbf{0.7613}    &{ }\textbf{0.7080}\\
&RF    &{ }\textbf{0.4520}    &{ }\textbf{0.7311}    &{ }\textbf{0.5050}    &{ }\textbf{0.4936}\\
&AdaBoost    &{ }\textbf{0.2795}    &{ }\textbf{0.0925}    &{ }\textbf{0.2842}    &{ }\textbf{0.2699}\\
\hline
   \Xhline{3\arrayrulewidth}
		\end{tabular}
	\end{center}
	\caption{The instance level Spearman ranked correlation between the indices and the prediction score increase of minority class sample over 80 real datasets. The highest correlation is shown in bold face.}
	\label{real_instance_table}
\end{table}%

\begin{table}[!t]
	\begin{center}
		\begin{tabular}{cccccc}
			    \Xhline{3\arrayrulewidth}
          &&OS&US&SMOTE&SW\\
\hline
\multirow{4}{*}{$kDN$}&SVM    &{ }0.4565    &{ }0.4531    &{ }0.4479    &{ }0.4607\\
&CART    &{ }0.4584    &{ }0.5742    &{ }0.5407    &{ }0.5052\\
&5NN    &{ }0.2738    &{ }0.3042    &{ }0.4527    &{ }0.3828\\
&RF    &{ }0.2792    &{ }0.5029    &{ }0.5597    &{ }0.1060\\
&AdaBoost    &{ }0.6820    &{ }0.7211    &{ }0.6499    &{ }0.5789\\
\hline
\multirow{4}{*}{$CL$}&SVM    &{ }0.2066    &{ }0.2695    &{ }0.1939    &{ }0.2010\\
&CART    &{ }0.2330    &{ }0.4520    &{ }0.3118    &{ }0.3037\\
&5NN    &{ }0.3736    &{ }0.3711    &{ }0.4473    &{ }0.3885\\
&RF    &{ }0.3497    &{ }0.4383    &{ }0.4769    &{ }0.2733\\
&AdaBoost    &{ }0.5474    &{ }0.4020    &{ }0.4020    &{ }0.5663\\
\hline
\multirow{4}{*}{$CM$}&SVM    &{ }0.1684    &{ }0.0304    &{ }0.1120    &{ }0.1774\\
&CART    &{ }0.0141    &{ }0.0935    &-0.0015    &{ }0.0619\\
&5NN    &{ }0.0420    &{ }0.0651    &{ }0.0343    &{ }0.1199\\
&RF    &{ }0.2167    &{ }0.1704    &{ }0.1603    &{ }0.1602\\
&AdaBoost    &{ }0.2913    &{ }0.2989    &{ }0.3425    &{ }0.2169\\
\hline
\multirow{4}{*}{$IR$}&SVM    &{ }0.2665    &{ }0.3744    &{ }0.3343    &{ }0.2629\\
&CART    &{ }0.3700    &{ }0.3151    &{ }0.4267    &{ }0.3414\\
&5NN    &{ }0.1492    &{ }0.1033    &{ }0.2843    &{ }0.1735\\
&RF    &-0.0500    &{ }0.1572    &{ }0.1905    &-0.1863\\
&AdaBoost    &{ }0.2656    &{ }0.2331    &{ }0.1781    &{ }0.2366\\
\hline
\multirow{4}{*}{$BI^3$}&SVM    &{ }\textbf{0.5423}    &{ }\textbf{0.5463}    &{ }\textbf{0.5395}    &{ }\textbf{0.5448}\\
&CART    &{ }\textbf{0.6314}    &{ }\textbf{0.6349}    &{ }\textbf{0.6854}    &{ }\textbf{0.6561}\\
&5NN    &{ }\textbf{0.4497}    &{ }\textbf{0.4406}    &{ }\textbf{0.6239}    &{ }\textbf{0.5497}\\
&RF    &{ }\textbf{0.3828}    &{ }\textbf{0.5420}    &{ }\textbf{0.6494}    &{ }\textbf{0.2035}\\
&AdaBoost    &{ }\textbf{0.7278}    &{ }\textbf{0.7693}    &{ }\textbf{0.7012}    &{ }\textbf{0.6249}\\
\hline
   \Xhline{3\arrayrulewidth}
		\end{tabular}
	\end{center}
	\caption{The data level Spearman ranked correlation between the indices and the improvement of F1 score by different imbalance recovery methods on data level over 80 real datasets. The highest correlation is shown in bold face.}
	\label{real_data_table}
\end{table}%

We specifically studied two real benchmark datasets from Table
\ref{real_data}: \textit{kddcup-land\_vs\_satan} and \textit{haberman}. The dataset \textit{kddcup-land\_vs\_satan} has $IR=75.67$ which is highly imbalanced and but $BI^3=0.02$, which means that the imbalance impact on this dataset is low. Table \ref{deeplook_1} shows the F1 score of different classifiers and the improvement of F1 score from the imbalance recovery methods. It can be observed that the F1 score for classifier without imbalance recovery is already very high. And therefore the improvements from the imbalance recovery methods are very limited. Most of them are close or equals to 0. US even deteriorate the F1 score for al classifier as shown negative improvement, which may be caused by that there is more decrease of precision than increase of recall as F1 is the harmonic mean between precision and recall. The result obtained from dataset \textit{kddcup-land\_vs\_satan} means the minority class in the dataset itself is very not difficult for classification even it is seriously outnumbered by the majority class. On contrary, The dataset \textit{haberman} has $IR=2.78$ which is not highly imbalanced compared with dataset \textit{kddcup-land\_vs\_satan}. But its $BI^3$ value is 0.2. Table \ref{deeplook_2} shows the F1 score and the improvements of different classifiers and imbalance recovery methods. It can be observed that most of the imbalance recovery methods can make obvious improvements on all classifiers. Most of the improvements of F1 score are greater than 0.1. Overall speaking, dataset $haberman$ is worthy for applying imbalance recovery methods because the F1 score can be actually improved , despite that its $IR$ is not very high. This example verifies again that $IR$ is not the only cause to the performance degeneration of imbalanced dataset.

\begin{table}[!tb]
	\begin{center}
		\begin{tabular}{cccccc}
			\Xhline{3\arrayrulewidth}
			&None & OS & US & SMOTE & SW\\\hline
			SVM&   0.9114& $+$0.0000 &   $-$0.5494 & $+$0.0000 & $+$0.0000\\
			CART&  0.9346 & $-$0.0050 &   $-$0.5495 & $-$0.0050 & $+$0.0000\\
			5NN&  0.9503 & $+$0.0000 &   $-$0.5906 & $+$0.0000 & $-$0.0169\\
			RF&  0.9446 &   $+$0.0358 &   $-$0.3950 &   $+$0.0356 & $+$0.0102\\
			AdaBoost&  0.9614 &   $+$0.0051 &   $-$0.5420 &   $+$0.0000 & $+$0.0000\\
			\Xhline{3\arrayrulewidth}
		\end{tabular}
	\end{center}
	\caption{The improvement of F1 score on the dataset \textit{kddcup-land\_vs\_satan}. The column None is the F1 score of the classifier without imbalance recovery methods.}
	\label{deeplook_1}
\end{table}%

\begin{table}[!tb]
	\begin{center}
		\begin{tabular}{cccccc}
			\Xhline{3\arrayrulewidth}
			&None & OS & US & SMOTE & SW\\\hline
			SVM&   0.0376& $+$0.1054 &   $+$0.4067 & $+$0.2108 & $+$0.1120\\
			CART&  0.3009 & $+$0.1130 &   $+$0.1386 & $+$0.0903 & $+$0.1452\\
			5NN&  0.2973 & $+$0.1201 &   $+$0.1270 & $+$0.1091 & $+$0.1025\\
			RF&  0.3514 &   $+$0.1676 &   $+$0.1813 &   $+$0.1482 & $+$0.1492\\
			AdaBoost & 0.3514 & $+$0.0533 &   $+$0.0659 &   $+$0.0671 & $+$0.0687\\

			\Xhline{3\arrayrulewidth}
		\end{tabular}
	\end{center}
	\caption{The improvement of F1 score on the dataset \textit{haberman}. The column None is the F1 score of the classifier without imbalance recovery methods.}
	\label{deeplook_2}
\end{table}%

\begin{figure}[!t]
	\centering
	\includegraphics[width=.7\linewidth]{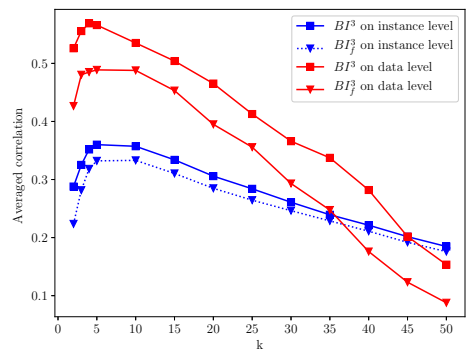}
	\caption{The change of correlation of $BI^3$ and $BI^3_f$ averaged over all classifiers and imbalance recovery methods as increasing the number of nearest neighbors $k$.}
	\label{select_k}
\end{figure}

The number of nearest neighbors $k$ used in calculation of $BI^3$ is set at 5 for all experiments. In this experiment ,we compare the averaged correlation of $BI^3$ with different settings of $k$. Besides, we also verify the effectiveness of the flexible $k$ that is adopted in Algorithm \ref{alg1}, compared with the one that just using the fixed number of $k$, which is denoted as $BI^3_f$. Figure \ref{select_k} shows the correlation of $BI^3$ averaged over all classifiers and imbalance recovery methods as increasing the number of nearest neighbors $k$ from 2 to 50. It can be observed that both instance level correlation and data level correlation have the highest value around $k=5$. As $k$ increases from 2 to 5, the averaged correlation increases and after that the averaged correlation decreases. That indicates the $k=5$ is a proper selection for $BI^3$. In addition, the averaged correlation of $BI^3$ is higher than $BI^3_f$ over all settings of $k$ for both data level and instance level correlation. That verifies the effectiveness of the flexible $k$.

\section{Concluding Remarks}

Most of the work presented in the area of class imbalance learning tries to recover the accuracy loss caused by imbalance ratio. However, the accuracy loss is related to not only imbalance but also many other data factors. Using IR to describe the classification difficulty of imbalance data is inaccurate and misleading. In this paper, we have proposed new measures $IBI^3$ and $BI^3$ to estimate the impact that is solely caused by imbalance on instance and data level, respectively. $IBI^3$ measures how much a single sample in the minority class is influence by the imbalance. $BI^3$, which is the average over $IBI^3$, can be used as a measure of degradation degree of imbalanced dataset, such that one can determine whether or not to apply imbalance recovery methods by referring to the value of $BI^3$ instead of IR. The experiments on synthetic and real benchmark datasets have shown high correlation on both instance level and data level with the improvements made by different imbalance recovery methods.

Along this work, there are still some rooms for the future work. For example, one work is to propose a classifier specific index, which shows exactly how much the imbalance influences a specific classifier, because each type of classifier has different sensitivity to imbalance. The second work is to incorporate $IBI^3$ into imbalance recovery methods, such as sampling or cost-sensitive methods, in order to help recovery the loss caused by imbalance. The third one is to take the advantages of $BI^3$ to guide the selection of a proper imbalance recovery method for a specific imbalanced data. Since recovery methods developed from the different theories and methodologies complement each other to a certain degree, their selection becomes especially important as given an imbalanced dataset.

\bibliographystyle{IEEEtran}
\bibliography{imba_ref_abv}




\end{document}